\documentclass[acmlarge]{acmart}

\usepackage{soul}
\usepackage{url}
\usepackage[utf8]{inputenc}
\usepackage{graphicx}
\usepackage{amsmath}
\usepackage{amsfonts}
\usepackage{amsthm}
\usepackage{booktabs}
\usepackage{algorithm}
\usepackage{algorithmic}
\usepackage{booktabs}
\usepackage{multirow}
\usepackage[utf8]{inputenc}
\usepackage{bbding}
\usepackage{tikz}

\newtheorem{defn}{Definition}

\newtheorem{assump}{Assumption}

\newcommand{\new}[1]{\textcolor{black}{#1}} % new content

\AtBeginDocument{%
  \providecommand\BibTeX{{%
    \normalfont B\kern-0.5em{\scshape i\kern-0.25em b}\kern-0.8em\TeX}}}

\setcopyright{acmcopyright}
\acmJournal{IMWUT}
\begin{document}

%%
%% The "title" command has an optional parameter,
%% allowing the author to define a "short title" to be used in page headers.
% \title{Privacy-preserving Deep Learning Model Customization for Edge Devices \\
\title{DistFL: Distribution-aware Federated Learning for Mobile Scenarios}

\author{Bingyan Liu}
\email{lby\_cs@pku.edu.cn}
% \authornote{The first two authors contributed equally.}
% \authornotemark[1]
\affiliation{%
  \institution{MOE Key Lab of HCST, Dept of Computer Science, School of EECS, Peking University}
%   \streetaddress{P.O. Box 1212}
  \city{Beijing}
  \state{China}
  \postcode{100871}
}

\author{Yifeng Cai}
% \authornotemark[1]
\email{caiyifeng@pku.edu.cn}
\affiliation{%
  \institution{MOE Key Lab of HCST, Dept of Computer Science, School of EECS, Peking University}
%   \streetaddress{P.O. Box 1212}
  \city{Beijing}
  \state{China}
  \postcode{100871}
}

\author{Ziqi Zhang}
\email{ziqi_zhang@pku.edu.cn}
\affiliation{%
  \institution{MOE Key Lab of HCST, Dept of Computer Science, School of EECS, Peking University}
%   \streetaddress{P.O. Box 1212}
  \city{Beijing}
  \state{China}
  \postcode{100871}
}

\author{Yuanchun Li}
\email{Yuanchun.Li@microsoft.com}
\affiliation{%
  \institution{Microsoft Research}
  \city{Beijing}
  \state{China}
  \postcode{100871}
}

\author{Leye Wang}
\email{leyewang@pku.edu.cn}
\affiliation{%
  \institution{Dept of Computer Science, School of EECS, Peking University}
%   \streetaddress{P.O. Box 1212}
  \city{Beijing}
  \state{China}
  \postcode{100871}
}

\author{Ding Li}
\email{ding_li@pku.edu.cn}
\affiliation{%
  \institution{MOE Key Lab of HCST, Dept of Computer Science, School of EECS, Peking University}
%   \streetaddress{P.O. Box 1212}
  \city{Beijing}
  \state{China}
  \postcode{100871}
}

\author{Yao Guo}
% \authornote{Correspondence to: Yao Guo.}
\email{yaoguo@pku.edu.cn}
\affiliation{%
  \institution{MOE Key Lab of HCST, Dept of Computer Science, School of EECS, Peking University}
%   \streetaddress{P.O. Box 1212}
  \city{Beijing}
  \state{China}
  \postcode{100871}
}

\author{Xiangqun Chen}
\email{cherry@pku.edu.cn}
\affiliation{%
  \institution{MOE Key Lab of HCST, Dept of Computer Science, School of EECS, Peking University}
%   \streetaddress{P.O. Box 1212}
  \city{Beijing}
  \state{China}
  \postcode{100871}
}

\renewcommand{\shortauthors}{Liu et al.}

%%
%% The "author" command and its associated commands are used to define
%% the authors and their affiliations.
%% Of note is the shared affiliation of the first two authors, and the
%% "authornote" and "authornotemark" commands
%% used to denote shared contribution to the research.

%%
%% The abstract is a short summary of the work to be presented in the
%% article.
\begin{abstract}
Federated learning (FL) has emerged as an effective solution to decentralized and privacy-preserving machine learning for mobile clients. While traditional FL has demonstrated its superiority, it ignores the non-iid (independently identically distributed) situation, which widely exists in mobile scenarios. Failing to handle non-iid situations could cause problems such as performance decreasing and possible attacks. Previous studies focus on the ``symptoms'' directly, as they try to improve the accuracy or detect possible attacks by adding extra steps to conventional FL models. However, previous techniques overlook the root causes for the ``symptoms'': blindly aggregating models with the non-iid distributions. In this paper, we try to fundamentally address the issue by decomposing the overall non-iid situation into several iid clusters and conducting aggregation in each cluster. Specifically, we propose \textbf{DistFL}, a novel framework to achieve automated and accurate \textbf{Dist}ribution-aware \textbf{F}ederated \textbf{L}earning in a cost-efficient way. DistFL achieves clustering via extracting and comparing the \textit{distribution knowledge} from the uploaded models. With this framework, we are able to generate multiple personalized models with distinctive distributions and assign them to the corresponding clients. Extensive experiments on mobile scenarios with popular model architectures have demonstrated the effectiveness of DistFL.
\end{abstract}

%% The code below is generated by the tool at http://dl.acm.org/ccs.cfm.
%% Please copy and paste the code instead of the example below.

% \ccsdesc[500]{Security and privacy~Privacy protections}

%%
%% Keywords. The author(s) should pick words that accurately describe
%% the work being presented. Separate the keywords with commas.

%%
%% This command processes the author and affiliation and title
%% information and builds the first part of the formatted document.
\maketitle
\section{Introduction}
\label{sec:intro}
% With the popularity of mobile devices and deep learning, more and more researchers are devoted to bring intelligence into these resource-limited devices for enriching human lives (e.g.,  smartphone-based speech assistant \cite{yang2019proxitalk}, sensor-enabled activity recognition \cite{chen2020metier,kwon2020imutube}). A typical solution to conduct mobile deep learning consists of two phases: (1) transmitting mobile data from devices to clouds \cite{han2016mcdnn} or edge servers \cite{li2018edge} in order to train the target model (e.g., DNN); (2) downloading the pre-trained model on devices for later inference. However, mobile data are collected by users and may contain some sensitive personal information, which will lead to privacy leakage if we directly upload these raw data. 

% Regulations on data privacy are getting more and more strict. For example, General Data Protection Regulation (GDPR) \cite{voigt2017eu} formally prohibits using data collected from clients arbitrarily. To comply strict regulations, \textit{federated learning} (FL) has been widely studied recently. FL allows training a machine learning model from a group of distributed clients without directly using 

As General Data Protection Regulation (GDPR) \cite{voigt2017eu} has been officially proposed to guarantee user privacy, it is no longer appropriate to transfer user data for cloud centralized training. Under this condition, \textit{federated learning} (FL), which enables a number of devices to collaboratively generate a global model without exposing their client data, has become an important and emerging research topic in recent years \cite{zhao2018federated,yang2018applied,yang2019federated}. A conventional FL solution consists of three  steps: (1) the server end distributes a model to each client \footnote{In the rest of this paper, we use \textit{client} to refer to \textit{mobile device}.}; (2) each client trains the offloaded model locally and uploads it to the server; (3) the server collects and coordinates these models by some aggregation algorithms (e.g., FedAvg \cite{mcmahan2017communication}) to generate a global model that owns the knowledge of each client. These three steps may repeat many times until the model converges.

\begin{figure}[t]
\centering
\includegraphics[width=0.8\columnwidth]{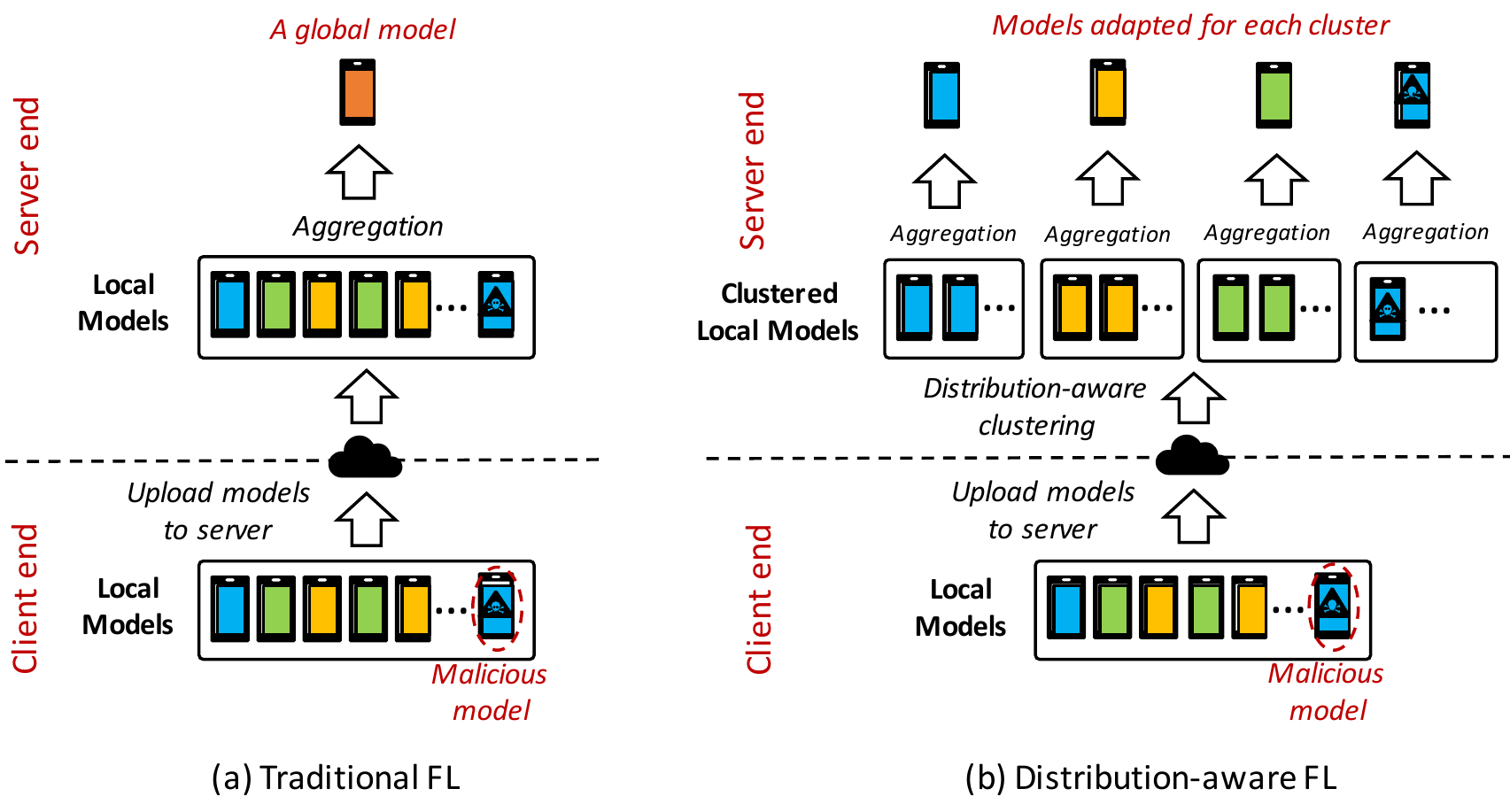} 
% \vskip -0.1in
\caption{Comparison between traditional FL and our distribution-aware FL. Here different colors represent different distributions.}
\label{fig:overview}
\end{figure}

Despite achieving tremendous success, conventional FL techniques may not be practical in certain scenarios that involve mobile devices. Existing FL techniques assume that the overall data distributions of all clients are iid (independently identically distributed). However, in mobile scenarios, this assumption may not always be valid. Data collected by a mobile device can be strongly biased on its context. For instance, pictures collected by a white-collar office worker may only contain indoor views while data from a traveller may be biased on outdoor landscapes. Simply averaging the models trained on these biased data sets may not achieve desired results.

Violating the iid assumption could be problematic for existing FL techniques in two ways. First, it may dramatically decrease the accuracy of trained machine learning models. In our experiment, conventional FL techniques, such as FedAvg~\cite{mcmahan2017communication}, can only achieve an accuracy of 21\% to 40\% with strongly biased data tasks. Second, in an even worse scenario, a malicious client can deliberately create a poisonous data set that misleads the global model to wrong predictions~\cite{tolpegin2020data,bagdasaryan2020backdoor}.

The research community has noticed the problem of violation of the iid assumption and many techniques have been proposed to fix the \textit{symptoms}. For example, some researchers propose to achieve personalization by manually conducting a loss-based clustering \cite{ghosh2020efficient} or a local adaptation after the FL process \cite{liang2020think}, which can improve the accuracy of traditional FL on non-iid data. People also propose to utilize model weights \cite{li2019abnormal,yu2020towards} or borrow extra data \cite{konstantinov2019robust} to assist the detection of malicious clients and remove them. Unfortunately, existing techniques suffer from limitations such as requiring massive manual tuning or lacking of sufficient accuracy.

We notice that the techniques mentioned in the above paragraph only focus on the ``symptoms'' rather than the root cause. Conventional FL techniques are inaccurate or vulnerable because data from clients are usually from different distributions while previous techniques still blindly aggregate models from these clients. They just specifically improve the accuracy or detect malicious clients by adding extra steps to the conventional FL process. However, these techniques fail to identify, distinguish and process clients with different data distributions separately, which is the fundamental cause of the symptoms of conventional FL techniques. 

Unfortunately, to the best of our knowledge, there is no existing technique that directly addresses the root cause of non-iid data distribution issues in FL. To fill the gap, in this paper, we try to fundamentally address the issue caused by the non-iid situation by proposing \textbf{distribution-aware FL}, a technique that clusters uploaded local models in terms of their \textit{distribution knowledge} and then aggregates them in each cluster. In this way, the overall non-iid situation can be approximately decomposed into several iid clusters and conventional FL techniques could be applied to each iid cluster. We believe this distribution-based clustering is possible because usually there are a large number of clients involved in the mobile FL system, some of which will inevitably come from the identical/similar data distribution and thus generate models with a similar distribution. For example, some mobile users may have a similar preference and thus collect data with a similar distribution.

Specifically, we propose \textbf{DistFL}, a framework that aims at achieving automated and accurate \textbf{Dist}ribution-aware \textbf{F}ederated \textbf{L}earning in a cost-efficient way. As shown in Figure \ref{fig:overview}, compared with traditional FL, our framework conducts a selective aggregation step based on the results of distribution-aware clustering. In this way, we are able to generate several personalized models with distinctive distributions and distribute them to corresponding clients with a similar distribution. Note that the malicious clients can also be clustered into a single group, eliminating their influence on normal clients and making FL more robust.

 Achieving distribution-aware FL is non-trivial. The key challenge is how to identify the data distribution of each client. Note that in FL, data on each client are private. DistFL cannot directly access these data. In other words, DistFL needs to infer the data distribution of each client based on the uploaded models. However, directly extracting input data distribution from a DNN could be hard as the DNN is usually viewed as a black box whose knowledge is uninterpretable. To address this challenge, we borrow the idea of synthesis technique~\cite{mahendran2015understanding}: we use the uploaded models to generate a set of bogus data and use it to measure the similarity of models. We say two models are similar if they generate similar classification results on the same data set, which is the generated bogus data set in our case. \new{Here the classification results refer to the probability distribution output at the last layer of the model, which is applicable to any DNN architectures.} By measuring the similarity between models, we can infer the data distribution that these models are generated from so that we can cluster models that are trained with similar data distributions. 

Clustering uploaded models with the inferred bogus data set requires overcoming three technical challenges. We list them as follows:
\begin{itemize}
    \item \textit{How to effectively generate bogus data that reflect the distribution of data from a model?} 
    A straight forward idea to generate bogus data is to directly use existing synthesis techniques. However, this simple idea suffers three main limitations: (i) existing synthesis techniques pay more attention to the label information, which can reflect the specific category features while may not benefit the clustering process on models with different distribution features; (ii) The synthesis process is computationally expensive especially when the number of labels is huge (e.g., The ImageNet dataset \cite{deng2009imagenet} has 1000 labels); (iii) Existing synthesis techniques focus on image tasks and do not try to generalize to other domains. In a word, it is not appropriate to use this method directly to generate representations of data distribution. 
    % It is intractable to obtain the distribution knowledge given the fact that we only have the access of uploaded models. First, the raw data cannot be uploaded and utilized due to privacy concerns, which invalidates the data-dependent information extraction. Second, the DNN is usually viewed as a black box whose knowledge is uninterpretable, which makes it harder to locate and extract the accurate distribution knowledge from the model itself.
    
    \item \textit{How to guarantee the efficiency during the generation process?} In real-world mobile scenarios the number of clients might be huge, which means that if we directly generate the distribution knowledge from each model, the computation costs may be unacceptable. So there is a high demand to design some optimization techniques in order to accomplish this process efficiently. This is a challenging problem considering the tremendous number of clients.
    
    \item \textit{How to automatically and accurately figure out the cluster strategy?} Unlike the previous work \cite{ghosh2020efficient} that accomplishes the clustering process by manually setting the number of clusters and using loss-based metric, we seek to an automated and precise method to decide the cluster information with the help of the extracted distribution knowledge.
    
\end{itemize}

DistFL addresses the first challenge mainly based on the following observation: \textit{The statistical information, which can represent the distribution knowledge \cite{liu2020pmc}, are exactly encoded in the widely used BatchNorm (BN) layers (i.e., running\_mean, running\_variance) of a DNN.} This finding inspires us to generate some "data" that can well fit the statistics of the BN layers. With the help of the synthesis technique \cite{mahendran2015understanding}, we are able to finally generate the "data" with distribution knowledge. To address the second challenge, we design two optimization methods to make the generation faster, ensuring the efficiency of the whole extraction process. To address the third challenge, we introduce a distribution-based clustering technique, where we feed the extracted "data" into each uploaded model and obtain the corresponding probability distribution vector at the Softmax layer. By calculating the KL divergence between them, we are able to generate a similarity matrix, which provides guidance for clustering by setting an apparent threshold, getting rid of manual tuning. In terms of the clustering results, we finally conduct aggregation inside each cluster and generate several personalized models for corresponding clients.

To summarize, DistFL consists of the following three steps: (1) a distribution extraction step to generate some data with distribution knowledge based on the uploaded models (see \S \ref{sec:knowledge_extract}); (2) a distribution-based clustering step to achieve automated and accurate clustering process by utilizing the generated distribution knowledge (see \S \ref{sec:knowledge_group}); (3) an aggregation step to selectively federate models with similar distributions (see \S \ref{sec:agg}). Besides, we would like to highlight that our framework does not need any raw data or feature information. Instead, we only operate the uploaded models in the server end, which indicates that our approach can be seamlessly integrated into the traditional FL pipeline to effectively cope with the existing non-iid situation.

% through extracting \textit{distribution knowledge} from uploaded models to represent the corresponding client property, FedExtract is able to conduct a knowledge-based grouping process. As a result, clients with similar distributions can be grouped accordingly and we can then implement the FL aggregation, generating a personalized model for each group. FedExtract has three main advantages: (1) The \textit{distribution knowledge} of the uploaded model can be efficiently and effectively extracted from BN layers that are widely used in modern DNN architectures; (2) The \textit{distribution knowledge} is a desirable property to distinguish clients with different data distributions, which contributes to accomplishing grouping automatically and accurately; (3) During the process, we do not need any raw data and only operate the uploaded models in the server end, which indicates that our approach can be seamlessly integrated into the typical FL pipeline. 

% In addition, FedExtract is also able to detect the malicious clients (e.g., injecting some ``poison'' to their local model for aggregation) and isolate them into a single group, eliminating their influence on normal clients and making FL more robust. We will detail the behavior of malicious clients and the limitation in Section \ref{sec:discussion}.

We evaluate DistFL on four simulated mobile scenarios since it is difficult to conduct FL in real-world mobile applications. The simulated scenarios include \textit{category-imbalance} scenario, \textit{environment-difference} scenario, \textit{privacy-protection} scenario and \textit{attack-injection} scenario, where we use public computer vision (CV) datasets and human activity recognition (HAR) datasets for concrete simulation (see \S \ref{sec:experiment_set}). Experimental results on these scenarios show that our proposed framework can reduce the relative error to other state-of-the-art personalization methods by up to 33.68\%. In addition, we are able to accurately detect the malicious clients and remove their influence, especially when the scale of malicious clients is large. For example, when there are 80\% attackers in the FL system, our approach can still achieve 8.05\% accuracy improvement and 7.33\% ASR (attack success rate) dropping compared to other defense methods, which validates the superiority of DistFL. Finally, we \new{conduct an experiment on a real-world dataset and} provide some in-depth analyses to further illustrate the effectiveness of the proposed framework. 

The contributions of this paper are summarized as three folds.
\begin{itemize}
    \item We propose \textit{distribution-aware federated learning}, a new paradigm to improve both the personalization and robustness performance of FL for mobile scenarios. By clustering similar models into iid groups and dealing with them separately, we can significantly mitigate the data heterogeneity and malicious clients influence caused by the non-iid situation.
    
    \item We design and implement DistFL to accomplish our goal, which achieves clustering through extracting and utilizing the distribution knowledge from the federated models. The whole process is efficient, automated and accurate. To the best of our knowledge, this is the first framework to explore and study the paradigm of distribution-aware federated learning.
    
    \item We conduct extensive experiments on multiple scenarios with different model architectures. The results demonstrate that the proposed DistFL performs better than state-of-the-art approaches.

\end{itemize}

\section{Related Work}
\label{sec:related-work}
This section presents a brief summary of existing literature related to our study.

\subsection{Mobile Federated Learning}

Federated learning (FL) is a prevalent distributed machine learning paradigm with improved privacy, which has been applied in many privacy-sensitive areas. For example, Google has developed the first product-level federated learning application \textit{Gboard} for next-word prediction \cite{hard2018federated}. NVIDIA has begun to cooperate with medical institutions aiming at building the first federated AI platform for medical diagnosis and drug researches \cite{nvdia_fl}. In addition, both industry and academic community have released some basic FL frameworks or benchmarks to support and contribute this promising field \cite{webank_fl,caldas2018leaf,he2020fedml,ryffel2018generic}. According to a recent survey \cite{li2019survey}, federated learning can be categorized into two typical types by the scale of federation:\textit{ cross-silo} and \textit{cross-device}. Here we mainly focus on the cross-device setting, where the participants are a large number of mobile or wearable devices rather than big organizations or data centers. In other words, this paper pays more attention to \textbf{mobile federated learning}, which has also been widely studied in recent years. For instance, Feng \textit{et al.} \cite{feng2020pmf} borrowed the idea of federated learning to predict human mobility. Samarakoon \textit{et al.} and Lu
\textit{et al.} \cite{samarakoon2019distributed,lu2019collaborative} designed a distributed federated learning system for connected vehicles. Guo \textit{et al.} \cite{guo2021prefer} proposed an edge-accelerated federated learning framework for POI recommendation. Next we describe some existing issues and investigate corresponding solutions in the context of mobile federated learning.

\textbf{Data Heterogeneity Problem and Personalization on Mobile FL.} A key problem in mobile FL is that the data on the users’ personal devices are usually non-iid (independently identically distributed) \cite{kairouz2019advances} due to different user environments or user preference. Faced with the challenge of statistical differences, a series of studies have been proposed to personalize the model. In particular, Zhao \textit{et al.} \cite{zhao2018federated} maintained a shared dataset across clients to improve training on non-iid data, which is impractical since it is intractable to find such a dataset and may cause privacy leakage. Liang \textit{et al.} \cite{liang2020think} used traditional FL to generate a global model and further fine-tune it with data in each local client. However, this method needs massive computation costs and cannot completely address the data heterogeneity problem since it still requires a traditional FL process (i.e., aggregating models with different distributions) before fine-tuning. Ghosh \textit{et al.} \cite{ghosh2020efficient} proposed the K-Cluster method to conduct FL, where they first set K clusters manually before different clients are respectively clustered and optimized in terms of the output loss. However, loss-based clustering is not accurate since the loss cannot reflect valuable distribution knowledge in the model. Moreover, this clustering algorithm needs to set the number of clusters manually before learning, which leads to big uncertainty because nobody has the ability to give a precise judgment to the number setting in advance. Different from these methods, DistFL partitions clients through inferring the data distribution directly from federated models, making the clustering process more efficient, automated and accurate.

% However, these methods cannot completely address the non-iid problem since the overall data distribution is not changed during their FL process. Different from them, our group-wise FL can manipulate the overall distribution by partitioning clients into different groups, transforming the non-iid condition into several iid data groups.

% Similar to group-wise FL, a recent work has proposed the idea of clustered FL \cite{ghosh2020efficient}, where different clients are respectively clustered and optimized. Specifically, this work proposed a method called \textit{K-Cluster} to estimate the cluster identities of each user in terms of the evaluation loss. However, loss-based clustering is not accurate since the loss cannot reflect valuable distribution knowledge in the model. Moreover, these clustering algorithms need to set the number of clusters manually before learning, which causes big uncertainty because nobody has the ability to give a precise judgment to the number setting in advance. Unlike the clustered FL solution, FedExtract partitions clients through extracting the \textit{distribution knowledge} from federated models, making the grouping process more efficient, automated and accurate.

% Dinh et al. \cite{dinh2020personalized} proposed pFedMe, where a new regularized loss function was designed to achieve better personalization.

% For example, Ghosh et al.\cite{ghosh2019robust} relied on a centralized clustering algorithm such as K-means to identify the cluster identities of all the users, resulting in high computational cost at the server. 

\textbf{Attack on Mobile FL.}
DNN attack is a hot and important research topic even before the emergence of FL \cite{liu2017trojaning,saha2020hidden,gu2019badnets}. The key idea is to inject some triggers inside the model to achieve malicious behaviors. Unlike existing attacks that target only one model in the centralized scenario, in mobile FL, the attack can happen in a single or several clients and is introduced on their local models to mislead the global model. We believe attacks on mobile FL are harder to defend because it is difficult to detect and locate the source of malicious behaviors considering the tremendous number of clients. 

A recent survey \cite{lyu2020threats} has summarized two types of attacks under the FL scenario: poison attacks \cite{tolpegin2020data,fung2018mitigating,biggio2012poisoning,bagdasaryan2020backdoor} and inference attacks \cite{hitaj2017deep,nasr2019comprehensive,melis2019exploiting,zhu2020deep}. The goal of poison attacks is to induce the FL model to output the target label specified by the adversary. For example, Tolpegin \textit{et al.} \cite{tolpegin2020data} implemented data poison attack by flipping the labels of training data from one class to another class in the local training epoch to mislead the global model output. Bagdasaryan \textit{et al.} \cite{bagdasaryan2020backdoor} conducted model poison attack by replacing the local model with elaborate weights, which can achieve a significantly higher attack success rate than data poison attacks. For inference attacks, model updates are utilized to infer an amount
of private information, such as class representatives \cite{hitaj2017deep}, membership as well as properties associated with a subset of the training data \cite{melis2019exploiting}. 

% In this paper, we mainly target how to defend the poison attacks under the FL system. For inference attacks, since it is 

% implement these attacks and observe the defense performance using our proposed approach.

\textbf{Defense on Mobile FL.}
Many efforts have been made to design defensive strategies to resolve attacks on mobile FL. Towards poison attacks, Konstantinov \& Lampert \cite{konstantinov2019robust} proposed the idea to maintain a small dataset in the server to provide extra guidance to mitigate the attack. However, the server end may not own such data and it is difficult to determine which data should be collected. Besides, Li \textit{et al.} \cite{li2019abnormal} proposed to use conv layers' weights of uploaded models to build an auto-encoder to detect malicious clients and Yu \& Wu \cite{yu2020towards} attempted to directly use the weights to distinguish  malicious clients for robust aggregation. Fu \textit{et al.} \cite{fu2019attack} applied a weighting scheme to give a low weight to uploaded attacked models, mitigating their negative influences. Different from these methods that devote to borrowing extra data or simply utilizing model weights, our approach 
isolates these malicious clients into a single cluster by detecting whether their \textit{distribution knowledge} is abnormal, which is more accurate and efficient.

For inference attacks, researchers applied differential privacy (DP) \cite{abadi2016deep} to federated learning by (1) clipping each client’s update, and (2) adding random noise \cite{geyer2017differentially,yu2020salvaging}. As a result, the privacy property of user data can be formally protected. In our experiments, we also implement DP on the client local model to defend the attack and test whether our approach is still effective under this \textit{privacy-protection} scenario.

\subsection{Knowledge Extraction from DNNs}
DistFL is motivated by the idea of synthesis while paying more attention to generate \textit{distribution knowledge} in an efficient manner. Generally, DistFL's process of generating bogus data can be categorized as the problem of extracting knowledge from DNN models. There are two lines of such work. The first is to use knowledge distillation (KD) \cite{hinton2015distilling}, where the model knowledge can be extracted/transferred to another model by mimicking the output or intermediate features. During the last few years, KD has been widely used in various fields, such as model compression \cite{liu2019wealthadapt,liu2021transtailor,mishra2017apprentice}, compact neural network architecture design \cite{romero2014fitnets} and semantic segmentation \cite{liu2019structured}. However, KD relies on the raw data to conduct training on the server end, which is unacceptable for FL due to privacy concerns. Besides, image synthesis \cite{mahendran2015understanding} can also be used to conduct knowledge extraction, where the model knowledge is derived to image-like data. For example, DeepDream \cite{mordvintsev2015inceptionism} and DeepInversion \cite{yin2020dreaming} tried to synthesize high fidelity and high resolution natural images from a trained DNN by gradient techniques. However, most of them focus on concrete label information and require massive computation costs.

\section{Preliminaries}
This section starts with a problem analysis and then explains the \textit{distribution knowledge} in detail. Finally, we formulate our objective.

\subsection{Problem Analysis}
\label{sec:problem_analy}
The aim of this subsection is to analyze how the data heterogeneity problem and the client malicious behaviors affect the FL process. Here we do not consider the inference attacks since they do not inject any malicious behaviors to the local model and thus will not harm FL. 

% At the beginning, we would like to point out that a recent work has shown that the typical FL aggregation algorithm (i.e., FedAvg) is ineffective when data is non-iid \cite{zhao2018federated}. 

For the data heterogeneity problem, each client owns a specific data distribution due to its environment or preference, which naturally satisfies the non-iid situation. For malicious behaviors in FL (i.e., poison attacks), they try to mislead specific data to wrong results, which can be considered to inject abnormal data information into the model. In other words, these attacks change the original data distribution of clients and also form a type of non-iid situation (normal and abnormal). So we can find that both of the issues come from aggregating models with non-iid distributions, which motivates us to address them fundamentally from a new perspective: \textit{Is it possible to decompose the non-iid situation into several iid situations by distribution-aware clustering?} If this clustering can be achieved, the data heterogeneity problem will disappear and models with malicious behaviors can be clustered into a single group, getting rid of their harmful influence on normal clients. In this paper, we design and implement distribution-aware FL to accomplish our goal.

% In addition to addressing the widely existed non-iid problem in FL systems, as shown in Figure \ref{fig:overview}, FedExtract can also detect the malicious clients that are attacked by injecting some data poison \cite{tolpegin2020data} and solely group them, making FL more robust. Here we discuss the reason \textit{why FedExtract can identify the poison attack}.

% First, we want to clarify that the key idea of data poison attack is to mislead special data to wrong results and inject abnormal data information into the model, which can be considered to change the original data distribution of clients. Therefore, FedExtract, which has the capability of detecting the model with abnormal distributions, can isolate these models into a single group, getting rid of their harmful influence. Besides, we would like to point out that FedExtract is not applicable to attacks that do not change the distribution of model parameters (e.g. inference attacks) and we assume that the server is honest and does not attempt to infer user privacy from the collaborative model.

\new{\subsection{Statistical Information and Distribution Knowledge}
\label{sec:dis_knowledge}}

% there are still two questions need to be considered: \textit{(1) which parts of the model can represent the distribution information? (2) how to generate the distribution knowledge in terms of these parts?} For the first question, as mentioned in Section \ref{sec:dis_knowledge}, distribution information can be characterized by the data-related \textit{mean} and \textit{var}, which are exactly encoded in the widely used BatchNorm (BN) layers (i.e., running\_mean, running\_variance) of the model. This finding inspires us to generate some "data" that can well fit the statistics in the BN layers. For the second question, 

\new{\textbf{Statistical Information.} Here the statistical information refers to a type of statistic that can well describe the overall distribution situation of a dataset. Usually the distribution parameters such as \textit{mean} and \textit{var} are used to represent the information \cite{liu2020pmc}. Note that these parameters are exactly encoded in the widely used BatchNorm (BN) layers (i.e., running\_mean, running\_variance) of a modern DNN model, which can be easily obtained in our scenario because all of the local models have been uploaded to the central server. }

\new{
\textbf{Distribution Knowledge.} In our work, we attempt to extract the \textit{distribution knowledge} from a DNN model to implement the later clustering. Intuitively, the distribution knowledge characterizes the situation of data distribution. In real-world applications, the same object image collected from different environments, the same activity coming from different people, the number of certain data based on different users, or some malicious behaviors injected by different attackers can be considered as typical examples that can lead to various distributions. In other words, the \textit{distribution knowledge} indicates that the specific feature information (e.g., the feature of the object image), which may be related to certain data, is identical while the overall data distribution situation (e.g., the environment that the object images are in) may be significantly different.
Therefore, we believe the distribution knowledge has a close relation to environments, special user preferences, and targeted behaviors, which is commonly seen in mobile scenarios.
}

\new{In order to extract the \textit{distribution knowledge}, we first denote it as a set of bogus data generated from an uploaded model that complies to the original distribution of the model's input data. As aforementioned, the original data distribution can be characterized by the statistical information (i.e., distribution parameters), which motivates us to generate data whose statistical information is similar to the original ones. Specifically, we employ the synthesis technique to optimize the noises to fit the distribution parameters for generating the \textit{distribution knowledge}. Details can be founded in Section \ref{sec:knowledge_extract}.}

% The main information DistFL uses to cluster uploaded models is the \textit{distribution knowledge}, which describes the data distribution from a model. Intuitively, the distribution knowledge is a set of bogus data generated from an uploaded model that complies to the original distribution of the model's input data.  

\subsection{Problem Formulation}

% \bingyan{we assume that the server is honest and does not attempt to infer user privacy from the collaborative model. delete because we have added differential privacy...}

We now introduce the symbols and annotations to formally define the optimization objective. Generally speaking, there are two parties involved in the FL process: \textit{client end} and \textit{server end}. The client end owns local datasets $D=\{D_1,D_2,...,D_N\}$ and use them to conduct training, generating local models $M=\{M_1,M2,...,M_N\}$. Here $N$ denotes the number of clients. The server end collects these models and aggregates them to generate a global model $M_{global}=Agg(M_1,M_2,...,M_N)$, where $Agg$ represents the aggregation algorithm (e.g., FedAvg). Unlike the traditional FL that outputs a global model, our objective is to generate several personalized models based on different distributions. In order to provide a formal definition of \textit{distribution-aware federated learning}, we first give an assumption.
\begin{assump}
\label{sec:assump}
Suppose $\mathcal{X}$ represents the data distribution. In the mobile FL system, there exist $Q$ types of data distributions $\mathcal{X}_1, \mathcal{X}_2,...,\mathcal{X}_Q$ and each local model $M_i \in M$ reflects one of them.  
\end{assump}

% \begin{assump}
% In the FL system, there always exist two local datasets $D_i$ and $D_j$ with similar or identical data distribution (i.e., $\mathcal{X}_i \approx \mathcal{X}_j$). Here $\mathcal{X}$ represents the data distribution.  
% \end{assump}

This assumption is commonly seen in the practical mobile FL scenario, where different distributions may come from diverse environments, user preference or injected malicious behaviors. Note that any client can find at least one client with similar distributions considering the huge number of clients in the mobile FL system and thus $Q < N$. 
Based on the assumption, we define \textit{distribution-aware federated learning} as follows.

\begin{defn}
\label{def:DFL}
\textbf{Distribution-aware Federated Learning (DistFL)}: Suppose clients can be ideally partitioned into Q clusters $(G_1,G_2,...,G_Q)$ in terms of their data distributions. At the aggregation stage, instead of aggregating all models as the traditional FL does, we conduct aggregation algorithms inside each cluster, generating Q personalized models $(M_p^1,M_p^2,...,M_p^Q)$ for clients, getting rid of the influence of the non-iid situation.
\end{defn}

To accomplish DistFL, we first construct a mapping $F: M_{target} \rightarrow K_{target}$, in order to map the target model $M_{target}$ to $K_{target}$, where $K_{target}$ is the distribution knowledge embedded in $M_{target}$. In terms of $K_{target}$, we attempt to characterize the relation among the models (clients) and allocate them into different clusters $(G_1,G_2,...,G_Q)$. In the following sections, we introduce a series of techniques to obtain $K_{target}$, the concrete cluster information $(G_1,G_2,...,G_Q)$ and the final personalized models $(M_p^1,M_p^2,...,M_p^Q)$.

% local model $M=\{M_1,M_2,...,M_N\}$ to $K=\{K_1,K_2,...,K_N\}$, where $K$ is the distribution knowledge embedded in $M$. In terms of $K$, we attempt to characterize the relation among the models (clients) and allocate them into different clusters $(G_1,G_2,...,G_Q)$. In the following sections, we introduce a series of techniques to obtain $K$, the concrete cluster information $(G_1,G_2,...,G_Q)$ and the final personalized models $(M_p^1,M_p^2,...,M_p^Q)$.

\begin{figure}[t]
\centering
\includegraphics[width=1\columnwidth]{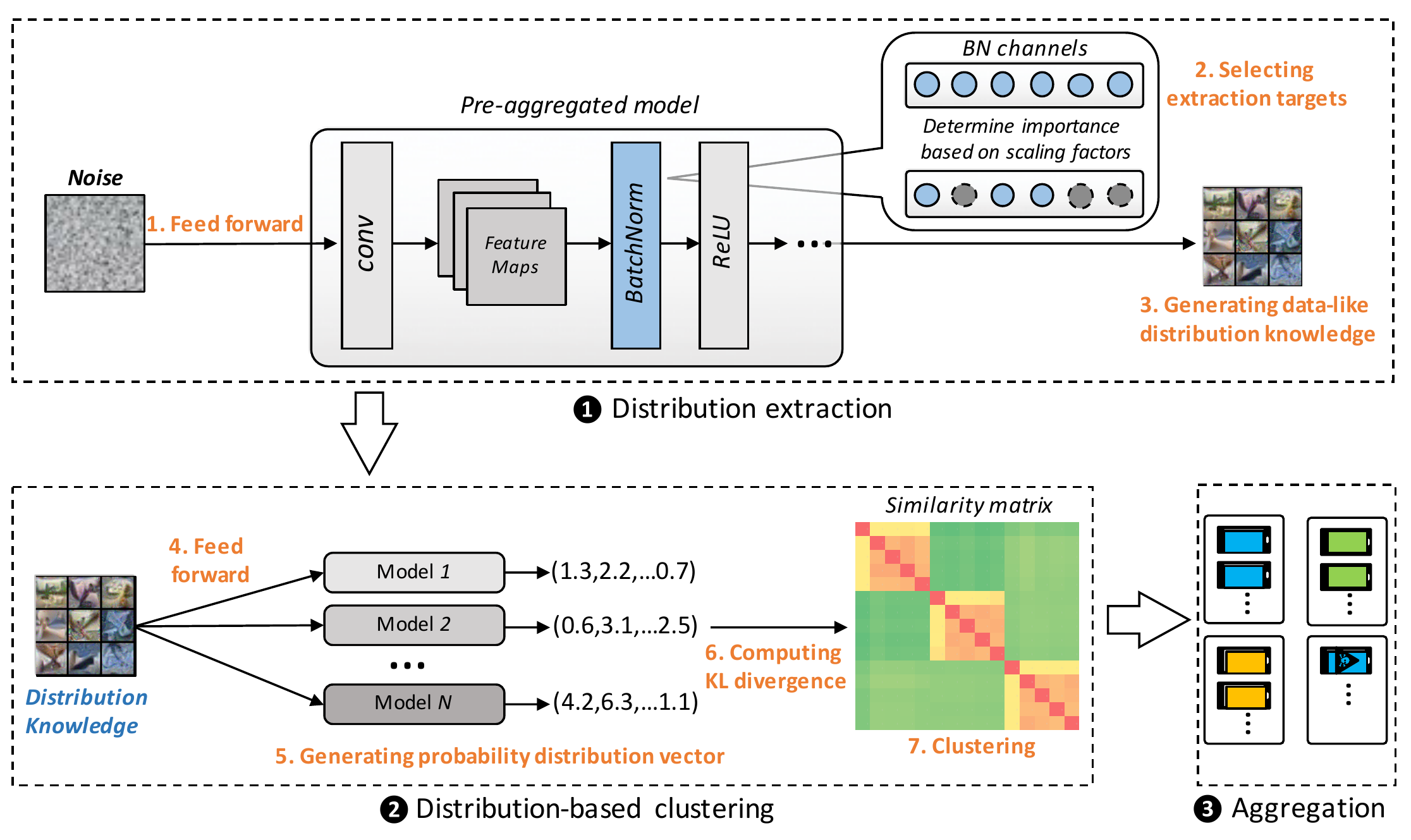} 
% \vskip -0.1in
\caption{Pipeline of our framework. The \textit{distribution extraction} step extracts the distribution knowledge from the target model and the \textit{distribution-based clustering} step leverages the extracted knowledge to cluster models with similar data distributions. Finally, the \textit{aggregation} step aggregates models inside each cluster, generating specialized models for each cluster.}
\label{fig:pipeline}
\end{figure}

% A recent work \cite{liu2020pmc} has proved that the distribution knowledge can be well described by feeding a dataset into a model and calculating the \textit{mean} and \textit{var} of the intermediate activation values. In other words, the data-dependent \textit{mean} and \textit{var} can represent the distribution knowledge, which motivates us to explore and exploit these statistical information from the uploaded models to obtain the knowledge. In the next sections, we will explain in detail how we manage to achieve the extraction and clustering based on the uploaded models only.

% Before introducing the concrete techniques, we first define what the distribution knowledge is and where we can extract them, which is the basis of our framework (see \S \ref{sec:dis_knowledge}). In terms of these concepts, we attempt to explore 
% how to extract and take advantage of the distribution knowledge for model clustering. Specifically, we partition uploaded models into different clusters with the assistance of the distribution knowledge. 
\section{The DistFL Framework}
\label{sec:approach}
This paper proposes an extraction-clustering based framework to achieve DistFL defined in Definition \ref{def:DFL}. In this section, we first give an overview of our framework and then describe each of the core components in detail.

\subsection{Overview}
We first discuss the overall process of DistFL.  

Figure \ref{fig:pipeline} illustrates the three key components in our framework, which we briefly summarize as follows:
\begin{itemize}
    \item \textbf{Distribution extraction}: We first need to extract distribution  knowledge for later clustering. Specifically, we randomly initialize a noise and feed it into an uploaded model(\textit{step 1}). Then in \textit{step 2} and \textit{step 3}, we select the extraction targets that may contain distribution information of the models and adapt the noise to data with distribution knowledge using backpropagation (i.e., synthesis techniques \cite{mordvintsev2015inceptionism,yin2020dreaming}). In order to make the process more efficient, we conduct a \textit{pre-aggregation} scheme to generate a global model and only extract the knowledge from it. Besides, the extraction targets of the model are decided by an \textit{importance-based target selection} method, which aims to remove some redundant information to save adaptation costs. As a result, we are able to extract the distribution knowledge of the target model (here $M_{target}=M_{global}$) and generate $K_{target}$.
    
    % we use image synthesis techniques \cite{mordvintsev2015inceptionism,yin2020dreaming} to adapt noises into image-like knowledge, with the help of the uploaded local model. To make the process more effective and efficient, we selectively extract the knowledge from important channels in BN layers, which can describe the distribution information and save massive computation costs.

    \item \textbf{Distribution-based clustering}: After obtaining the distribution knowledge $K_{target}$, we feed them into each uploaded model to observe their response (\textit{step 4}). Concretely, we generate the probability distribution vector at the Softmax layer (\textit{step 5}). By calculating the KL divergence \cite{van2014renyi} between these vectors, we are able to construct a similarity matrix, which provides guidance to cluster models without manual efforts (\textit{step 6} and \textit{step 7}). In this way, the cluster information $(G_1,G_2,...,G_Q)$ is formed.
    
     \item \textbf{Aggregation}: Instead of blindly aggregating all of the uploaded models as traditional FL does, our approach conducts a selective aggregation process by only aggregating models inside the same cluster, generating corresponding personalized models $(M_p^1,M_p^2,...,M_p^Q)$ for clients with different data distributions.
\end{itemize}

We dynamically implement these steps in each round of FL until convergence. Note that the whole process is operated in the server, without consuming extra resources on the resource-constrained clients. 

\subsection{Distribution Extraction}
\label{sec:knowledge_extract}
The first step is to extract the distribution knowledge from uploaded models. Given an uploaded model, DistFL generates the bogus input that represents the input data distribution of the model. To achieve this, we borrow the idea of the image synthesis technique \cite{mordvintsev2015inceptionism}, whose goal is to synthesize high fidelity and high resolution natural images given a single model, which can be regarded as the knowledge of the model. \new{Here we do not use GAN \cite{goodfellow2014generative} because it requires not only the model but also the user raw data to generate bogus data, which is unrealistic in our scenario.}  

Specifically, given a randomly initialized noise $\hat{x} \in \mathcal{R}^{C\times H \times W}$, and a target label $y$, the synthesis technique is able to generate label-related images (knowledge) by optimizing the following objective
\begin{equation}
\label{eq:image_synthesis}
    \min_{x^r} \mathcal{L}(\hat{x},y) + R(\hat{x})
\end{equation}
where C, H, and W are the number of channels, height and width of the noise. $x^r$ is the reconstructed image. $\mathcal{L}(\cdot, \cdot)$ denotes the loss function and $R(\cdot)$ represents an image regularization term to improve the visual quality.

In our work, different from previous works that devote to generating high fidelity natural images with target labels, we focus on optimizing the noises to fit the statistics encoded in the BatchNorm (BN) layers (i.e., running\_mean, running\_variance) of the uploaded model as much as possible, no matter whether the generated items are natural images or not. Therefore, our approach can be applied to any tasks by changing the dimension of the initialized noise. For instance, to the task of human activity recognition (HAR), the noise is a 1-dimension vector, which can also achieve good performance as shown in our experiments (see \ref{sec:environment-difference}). Here we still take the image task as an example for a better comparison to traditional synthesis techniques.

Concretely, given a series of noises $\hat{X}=\{\hat{x_1},\hat{x_2},...,\hat{x_z}\}$ ($z$ represents the number of noises) and the target model $M_{target}$, we attempt to address an optimization problem as follows
\begin{equation}
\label{eq:optim_bn}
\begin{aligned}
    \min_{K_{target}} \sum_{i=1}^H \Big( ||\mu_i(\hat{X})-BN_i(running\_mean)||_2  +  ||\sigma^2_i(\hat{X})-BN_i(running\_variance)||_2 \Big)
\end{aligned}
\end{equation}
where $K_{target}=\{k_{target}^1,k_{target}^2,...,k_{target}^z\}$ is the reconstructed distribution knowledge from the target model $M_{target}$. $\mu_i(\hat{X})$ and $\sigma^2_i(\hat{X})$ denote the mean and variance estimates of feature maps to the $i_{th}$ layer, which are generated by feeding the noises into the model for an inference process. $H$ represents the number of BN layers of the model. $BN_i(running\_mean)$ and $BN_i(running\_variance)$ are the parameters in the $i_{th}$ BN layer, which can reflect the statistical information. $||\cdot||_2$ represents the $L_2 \ norm$. After solving this optimization problem by gradient-descent based back propagation, we can obtain the final distribution knowledge $K_{target}$.

\textbf{Efficiency optimization 1.} 
The naive approach of knowledge extraction is to generate the distribution knowledge for each of the uploaded models respectively. However, in an FL system, we may have a huge number of clients, each of which holds a local model. Thus directly extracting the distribution knowledge from all uploaded models could be time-consuming and computationally expensive. To improve the efficiency of knowledge extraction, instead of extracting the distribution from all of the uploaded models, we  apply a \textit{pre-aggregation} scheme to generate a global model and use it to extract the distribution knowledge. The intuition is that no matter what changes the attackers make to their local models, the aggregation process must be enabled. Otherwise, FL cannot be continued, let alone injecting malicious behaviors. Besides, the aggregated model must be a usable model for later deployment, indicating that we are able to extract the distribution knowledge from it although the distribution may be abnormal.

Specifically, we first generate the global model $M_{global}$ by averaging all of the uploaded models as the FedAvg algorithm does \cite{mcmahan2017communication}:
\begin{equation}
\label{eq:pre-agg}
\begin{aligned}
   M_{global}=\frac{1}{N} \sum_{i=1}^N M_i
\end{aligned}
\end{equation}
The generated $M_{global}$ can be regarded as the target model $M_{target}$.

\textbf{Efficiency optimization 2.} As DNNs become larger and larger for processing more complex tasks, the number of BN layers have also increased dramatically. Under this situation, selecting all of the BN layers to conduct aforementioned optimization may cause considerable computation costs. Therefore, to make the process more efficient, we present an \textit{importance-based target selection} method to selectively extract the \textit{important} channels of BN layers to implement the optimization. Here the importance is measured by the \textit{scaling factor} attached in each BN channel and we pick out $E\%$ ( \new{ i.e., the ratio of selected channels to the whole channels of the BN layer}) channels with the larger factor value in each layer. Thus, the optimization problem is reformulated as 
\begin{equation}
\label{eq:efficient_optim}
\begin{aligned}
    \min_{K_{target}} \sum_{i=1}^H \Big( ||\mu_i(\hat{X})-BN_i^S(running\_mean)||_2  +  ||\sigma^2_i(\hat{X})-BN_i^S(running\_variance)||_2 \Big),
\end{aligned}
\end{equation}
where $S$ represents the selected BN channel set in the corresponding layer. 
% We defer the readers to Section \ref{xx} for more details of the efficiency analysis. 

In terms of Equation \ref{eq:efficient_optim}, we are able to generate the distribution knowledge of the target model $M_{target}$ and obtain the final $K_{target}$.

% Different from previous work that devote to generating high fidelity natural images, we only feed and adapt the noises to fit the BN parameters as much as possible, no matter whether the generated items are natural images. The intuition is that the parameters in BN layers comprise some statistical information, which may have close relation to the raw data distribution. Therefore, we believe the generated items can well describe the distribution knowledge. Besides, to make the process more efficient, we selectively extract the important channels of BN layers and only execute the process for one model.

\subsection{Distribution-based Clustering}
\label{sec:knowledge_group}
This component determines the concrete cluster information $(G_1,G_2,...,G_Q)$ based on the extracted $K_{target}$. Because the extracted knowledge is in a form of "data", we can directly feed them into each uploaded model to observe the responses, which can then be used to guide clustering.

% how to aggregate to get the global model and invert the distribution knowledge $K_{global}$

Our intuition is that, given the same distribution knowledge, models with similar distributions will give similar responses. Therefore, clustering models has the same effect as clustering their corresponding responses, which simplifies our goal. Specifically, given an uploaded model $M_c$ and the target distribution knowledge $K_{target}$, we feed each input $k_{target}^i$ into the model and concatenate the probability distribution vector at the final Softmax layer
\begin{equation}
\label{eq:inference}
\begin{aligned}
   V_c=concat \{M_c(k_{target}^i)\}_{i=1,2,...,z}
\end{aligned}
\end{equation}
In this way, we generate $V_1,V_2,...,V_N$ from each uploaded model to represent its response to the certain distribution knowledge (i.e., $K_{target}$). \new{Notice that the responses generated from each model are a series of probability distribution vectors, and thus we can calculate their similarity with the KL divergence \cite{kullback1951information}, which has been proved effective in measuring the difference degree between two probability distributions.} Next, we calculate the KL divergence between these vectors as follows
\begin{equation}
\label{eq:sim}
\begin{aligned}
   Div(p,q)=D_{KL}(V_p||V_q)
\end{aligned}
\end{equation}
Based on Equation \ref{eq:sim}, we can generate a similarity matrix $SIM$, where $SIM_{pq}$ represents the distribution similarity degree for the $p_{th}$ model to the $q_{th}$ model and the lower value implies higher similarity. In terms of $SIM$, we are able to easily determine the number of clusters and which model is in which cluster (i.e., $(G_1,G_2,...,G_Q)$) by setting a threshold (the threshold is apparent as shown in the right part of Fig \ref{fig:noise_konwledge}, \new{where every 5 clients with similar distributions have a significantly higher similarity compared to others and thus can be clearly clustered}), getting rid of the complex and inaccurate manual efforts.Specifically, given an uploaded model, we set a threshold to pick out the similar models based on $SIM$ to form a rough iid cluster. For other unselected models, we follow the above step and form another iid cluster. In this way, we can finally partition these models into different iid clusters without manual setting of the concrete number of clusters. 

Besides, we would like to point out that although we do not know the concrete information of the dataset or problem settings, the manual threshold is always effective because the computed matrix $SIM$ has clearly described the difference between models generated from iid data distributions and non-iid data distributions, which benefits the threshold-based partition. In other words, the value of $SIM$ is close in the iid group while varying greatly between two groups (i.e., iid and non-iid). Therefore, the threshold can be always and easily generated in terms of the distance among the $SIM$ values and we finally form different model groups based on it. In summary, what makes the threshold works is that the uploaded models intrinsically come from different groups according to our Assumption \ref{sec:assump} and the matrix $SIM$ is able to characterize their similarity, which make the threshold feasible in partitioning them.

% Here we would like to highlight that although the threshold in our approach needs to be set manually, it is easy to find a suitable distinguishing value since the similarity degree between models generated by iid distributions and non-iid distributions is extremely different, which, to some extent, can be considered as an automated process.}

% Finally, we conduct FedAvg \cite{mcmahan2017communication} in each group and generate our desirable personalized models $(M_p^1,M_p^2,...,M_p^Q)$.

\subsection{Aggregation}
\label{sec:agg}
Model aggregation is the final step to finish FL. A prevalent and effective aggregation is FedAvg \cite{mcmahan2017communication}, which has been proved effective especially when the data is iid. Here we adopt this aggregation algorithm because we have partitioned uploaded models into several iid clusters. Concretely, for the cluster $G_i$, we average the models inside this cluster (i.e., $M_{G_i}^1,M_{G_i}^2,...,M_{G_i}^{N_i}$) and generate the corresponding personalized model $M_p^i$
\begin{equation}
\label{eq:final-avg}
\begin{aligned}
   M_p^i=\frac{1}{N_i} \sum_{j=1}^{N_i} M_{G_i}^j ,
\end{aligned}
\end{equation}
where $N_i$ is the number of models inside the cluster $G_i$. In this way, we are able to obtain the final personalized models $(M_p^1,M_p^2,...,M_p^Q)$ and deploy them to corresponding clients.

\new{\subsection{Theoretical Analysis}
\label{sec:theory}}
In this subsection, we provide theoretical analysis to support our framework. A possible direction is to quantify the uncertainty of models built by different aggregation algorithms based on the theory proposed by Gal \cite{kendall2017uncertainties}. After exploration, we find that the key factor to affect model uncertainty is the amount of data. In other words, the more data we have, the smaller uncertainty the model holds. In our FL scenario, the client data is not changed when applying different aggregation algorithms and thus we cannot explain the improved performance from this perspective. Here we use the \textit{weight divergence} to provide feasible theoretical analysis. 

Specifically, we analyze the \textit{weight divergence} among the candidate models that need to be aggregated in traditional FL and DistFL, which has been proven to be a key factor to the final FL performance (i.e., higher divergence leads to worse performance \cite{zhao2018federated}). According to the work \cite{zhao2018federated}, the \textit{weight divergence} satisfies the following inequality
\begin{equation}
\begin{aligned}
\left\|\boldsymbol{w}_{m T}^{(f)}-\boldsymbol{w}_{m T}^{(c)}\right\| \leq & \sum_{k=1}^{N} \frac{n^{(k)}}{\sum_{k=1}^{N} n^{(k)}}\left(a^{(k)}\right)^{T}\left\|\boldsymbol{w}_{(m-1) T}^{(f)}-\boldsymbol{w}_{(m-1) T}^{(c)}\right\| \\
&+\eta \sum_{k=1}^{N} \frac{n^{(k)}}{\sum_{k=1}^{N} n^{(k)}} \sum_{i=1}^{C}\left\|pro^{(k)}(y=i)-pro(y=i)\right\| \sum_{j=1}^{T-1}\left(a^{(k)}\right)^{j} g_{\max }\left(\boldsymbol{w}_{m T-1-k}^{(c)}\right)
\end{aligned}
\end{equation}
where $N$ is the number of clients. $\boldsymbol{w_{m T}}$ denotes the model weight at the $mT$ round. $f$ and $c$ represent the federated setting and the centralized setting. $C$ is the number of category and $y$ is the ground truth. $g_{max}$ is the max gradient and $a$ is a hyper-parameter. $pro()$ represents the probability when a specific condition is satisfied. 

Here we mainly focus on how to compare the candidate models' \textit{weight divergence} between traditional FL and our DistFL. Based on the inequality, the distribution heterogeneity ( i.e., $\sum_{i=1}^{C}\left\|pro^{(k)}(y=i)-pro(y=i)\right\|$ in the inequality) plays a key role in the divergence degree: higher heterogeneity leads to larger divergence. In traditional FL, the distribution heterogeneity is high since the data distribution under the candidate models is typically non-iid. Different from it, DistFL partitions uploaded models into several iid clusters to conduct aggregation, which significantly mitigates the heterogeneity and decreases the weight divergence degree. Therefore, compared with traditional FL, DistFL maintains a low weight divergence degree during aggregation, contributing to improved performance.

\section{Evaluation}
\label{sec:experiments}

This section presents the evaluation of DistFL over different mobile scenarios and compares it against various baseline methods. Specifically, we primarily look at the following two aspects:
\begin{itemize}
    \item What is the performance of DistFL on our simulated mobile scenarios \new{and the real-world scenario}? How does it compared to existing methods? (\S \ref{sec:category-imbalance}, \S \ref{sec:environment-difference}, \S \ref{sec:privacy-enhance}, \S \ref{sec:attack-inject}, \S \ref{sec:real}, \S \ref{sec:convergence})
    
    \item Is the extracted distribution knowledge effective in distinguishing models with different distributions? How to select parameters for better efficiency during the extraction process?(\S \ref{sec:detailed_analy})
    
    % \item How does our efficiency optimization affect the effectiveness of our framework? (\S \ref{})
\end{itemize}

% In the following, we first describe the settings of our experiments and then evaluate our framework based on the above questions.

\begin{table}[]
\centering
\caption{The simulation settings of different datasets. Note that the \textit{HAR-ADL} dataset is collected from real mobile users, which means that it does not have the specific distribution type.}
\label{tab:simulated_dataset}
% \vskip 0.15in
\begin{tabular}{lc|c|c|c}
\hline
\multicolumn{1}{c|}{\textbf{Dataset}}     &
\multicolumn{1}{c|}{\textbf{Distribution type}}&
\multicolumn{1}{c|}{\textbf{Client index}} & \multicolumn{1}{c|}{\textbf{\#Training sample}} & \multicolumn{1}{c}{\textbf{\#Testing sample}}
     \\ \hline
\multicolumn{1}{c|}{\multirow{5}{*}{\textit{CIFAR-10}}}      & \textit{Type1} &   1,2,...,20                           & 800                                   & 2000                                  \\
\multicolumn{1}{l|}{}                                     & \textit{Type2}&   21,22,...,40                          & 800                                   & 2000                                  \\
\multicolumn{1}{l|}{}                                       & \textit{Type3}  &      41,42,...,60                      & 800                                   & 2000                                 \\
\multicolumn{1}{l|}{}                                       & \textit{Type4}   &        61,62,...,80                      & 800                                   & 2000                                  \\
 \multicolumn{1}{l|}{}                                     & \textit{Type5}  &        81,82,...,100                      & 800                                   & 2000 \\ \hline
\multicolumn{1}{c|}{\multirow{4}{*}{\textit{Office-Home}}}   & \textit{Ar}   & 1,2,...,5                           & 1940                                  & 485                                   \\
\multicolumn{1}{c|}{}                                      & \textit{Cl}   & 6,7,...,10                          & 3490                                   & 873                                  \\
\multicolumn{1}{c|}{}                                     & \textit{Pr}    & 11,12,...,15                      & 3550                                   & 887                                  \\
\multicolumn{1}{c|}{}                                      & \textit{Rw}    & 16,17,...,20                      & 3485                                  & 871                                  \\ \hline
\multicolumn{1}{c|}{\textit{HAR-ADL}}  & \textit{unknown}  & 1,2,...,20 & 5408 & 1984 \\ \hline
\end{tabular}
% \vskip -0.1in
\end{table}

\subsection{Experimental Settings}
\label{sec:experiment_set}
In the following, we will describe the simulation settings of our experiments since it is hard to conduct experiments on real-world mobile applications. The source code is available at \url{https://github.com/YikFungTsai/DistFL}.

\textbf{Used public datasets.} 
We mainly use  public computer vision (CV) datasets and human activity recognition (HAR) datasets for simulation. Concretely, we select CIFAR-10 \cite{krizhevsky2009learning}, Office-Home \cite{venkateswara2017deep} and HAR-ADL \cite{anguita2013public} to simulate various situations. The detailed usage and processing of these datasets are explained in the next part.

\textbf{Simulated mobile scenarios.}
We design four commonly seen mobile scenarios based on public datasets, whose detailed simulation settings are illustrated in Table \ref{tab:simulated_dataset}. Here we briefly summarize each scenario as follows.
\begin{enumerate}
    \item \textit{Category-imbalance scenario.} This scenario means that different clients may hold different categories  due to the user preference. For example, one client may own many "dog" images while another prefers to collect "cat" images. Similar to the previous work \cite{mcmahan2017communication}, we simulate this situation by the public CIFAR-10 dataset, which contains 10 different image categories. Considering that the client end may own limited data, we uniformly select 4000 samples from the training set and allocate them into 100 clients, where every 20 clients belong to a type of distribution. Therefore, in total we have 5 types of distribution, each of which holds two disjoint categories of CIFAR-10. For testing, we use the official testing set and partition them into 5 parts based on the 5 types of distribution. Here we use \textit{TypeX (X=1,2,...,5)} to denote each type of distribution.
    
    \item \textit{Environment-difference scenario.} This scenario suggests that the main recognition object is identical while the background is different due to diverse environments. Under this circumstance, we use the Office-Home dataset for simulation, which contains 15,500 images with four distinctive domains: Artistic images (Ar), Clipart images (Cl), Product images (Pr) and Real-World images (Rw). Each domain has 65 identical object categories but different backgrounds. As a recent work does \cite{liu2021pfa}, we first partition each domain into a training set (80\%) and a testing set (20\%) since this dataset has no official train/test split. For training sets, we further divide each domain into 5 parts and each client owns one part. Therefore, in total we have 20 clients with 4 types of data distribution and each type is evaluated on the testing set of the corresponding domain. 
    
    \item \textit{Privacy-protection scenario.} Considering the inference attack may steal some sensitive user information from the uploaded models or gradient updates, we apply differential privacy (DP) for the local training to defend this type of attack as prior works do \cite{geyer2017differentially,yu2020salvaging}. Under this privacy-protection scenario, we want to examine whether our approach can still work given the DP-based local models. Here we take Office-Home as an example to observe the performance on different privacy budgets.
    
    \item \textit{Attack-injection scenario.} This scenario indicates that there exist some attackers who inject some malicious behaviors to their local models to mislead FL. Here we mainly focus on the poison attack in the context of FL because the inference attack can be effectively defended if we adopt DP techniques. Specifically, we implement the label flipping attack \cite{tolpegin2020data} and the model replacement attack \cite{bagdasaryan2020backdoor} to some clients for simulation. Notice that many baseline methods do not target the non-iid problem, thus we only simulate the iid setting on CIFAR-10 (i.e., 50 clients, each of which has 10 categories with 1000 samples) to make a fair comparison.
\end{enumerate}
Besides, we use the HAR-ADL \cite{anguita2013public} dataset to simulate both the category-imbalance and the environment-difference scenario, which is collected from accelerometers and gyroscopes of 30 smartphone users who perform 6 different activities (\textit{i.e., walking, walking\_upstairs, walking\_downstairs, sitting, standing, laying)}. Because HAR-ADL is a real world dataset, where different users may have specific user preference under diverse environments, we believe it can cover both of the first two scenarios. Specifically, we select 20 users as 20 clients and use a simple 2-layer CNN to conduct experiments.

% since the dataset is collected from real users and is not specially designed for any specific distribution property. We believe this is more practical for mobile scenarios. Specifically, we \ref{xx} \bingyan{select 20 to keep roughly consistent to other datasets.}

% \textbf{HAR-ADL}. HAR-ADL is a dataset gathered from accelerometers and gyroscopes of smartphones from 30 volunteers performing 6 different activities \textit{(walking, walking\_upstairs, walking\_downstairs, sitting, standing, laying)}. The volunteers followed a protocol which lasts 192 seconds, performing it twice, with a Samsung S II mounted on the left side of a belt around the waist for the first time, and the same phone placed on a location on the belt where the user preferred during the second time. Sensor signals from the accelerometer and the gyroscope were collected at 50 Hz. Here we randomly select 20 subjects as 20 clients, and we use a simple 2-layer CNN with Batch Normalization in experiments.

\textbf{Models.}
We use three types of model architectures: simple 2-layer CNN, VGGNet \cite{simonyan2014very} and ResNet \cite{he2016deep}, to test the performance of our approach. Towards VGGNet, we replace the original fully connected layers with a global average pooling layer in order to alleviate overfitting \cite{qiao2019neural}. 

% Specifically, different models are used to different scenarios: VGG-11 and ResNet-18 on the \textit{category-imbalance scenario}, ResNet-50 on the 

\textbf{Compared baseline methods.}
We compare the proposed approach to two lines of work. The first one is the personalization methods targeting at the non-iid problem in FL systems. Specifically, the following methods are compared:
\begin{itemize}
    \item \textit{Local}: This method means that we only train a model at each local client without federation.
    
    \item \textit{LG-Net} \cite{liang2020think}: LG-Net first learns a global model by traditional FL and then fine-tunes the model with the data in each client.
    
    \item \textit{K-Cluster} \cite{ghosh2020efficient}: Similar to our approach, K-Cluster also aims to cluster the uploaded models and conduct aggregation accordingly. However, this method is only based on simple loss values and requires setting the cluster manually, which is inaccurate and impractical since it is hard to know the cluster information before learning.
\end{itemize}

% \textit{Local}, \textit{LG-Net} \cite{liang2020think} and \textit{K-Cluster} \cite{ghosh2020efficient}. Here \textit{Local} means that we only train a model at each local client without federation. Others are personalization methods designed for FL. 

Another line of work is the defense methods, which aim to resist poison attacks for FL systems. Concretely, we implement three methods for comparison: 
\begin{itemize}
    \item \textit{Credit} \cite{li2019abnormal}: Credit attempts to use conv layers' weights of uploaded models to build an auto-encoder such that the abnormal clients can be selected out.
    
    \item \textit{IRLS} \cite{fu2019attack}: IRLS applies a weighting scheme that gives a lower weight to uploaded attacked models, thus mitigating their negative influence.
    
    \item \textit{GRA} \cite{yu2020towards}: The goal of GRA is to directly use the weights to distinguish malicious clients for robust aggregation.
\end{itemize}
Besides, the typical \textit{FedAvg} \cite{mcmahan2017communication} is also compared as a baseline for both of the two lines.

\textbf{Implementation details.}
We performed all experiments with the PyTorch framework, based on a server that has 4 GeForce GTX 2080Ti GPUs, 48 Intel Xeon CPUs, and 128GB memory. In the \textit{category-imbalance scenario}, we tested the performance on VGG-11 and ResNet-18 with the CIFAR-10 dataset, where input images are 32*32 and normalized to zero mean for each channel. We assigned the learning rate to 1e-4 for VGG-11 and 1e-2 for ResNet-18 respectively. In the \textit{background-difference scenario}, we used the pre-trained ResNet-50 model to test the performance on the Office-Home dataset, where we randomly cropped the input images to 224*224 and these images are also normalized to zero mean for each channel. The learning rate was set to 1e-2. Besides, we conducted experiments in the HAR-ADL dataset with a simple 2-layer CNN, where we sampled data at every 50 Hz and set the learning rate to 1e-2.

% \textbf{HAR-ADL}. HAR-ADL is a dataset gathered from accelerometers and gyroscopes of smartphones from 30 volunteers performing 6 different activities \textit{(walking, walking\_upstairs, walking\_downstairs, sitting, standing, laying)}. The volunteers followed a protocol which lasts 192 seconds, performing it twice, with a Samsung S II mounted on the left side of a belt around the waist for the first time, and the same phone placed on a location on the belt where the user preferred during the second time. Sensor signals from the accelerometer and the gyroscope were collected at 50 Hz. Here we randomly select 20 subjects as 20 clients, and we use a simple 2-layer CNN with Batch Normalization in experiments.

Towards the \textit{privacy-protection scenario} and the \textit{attack-injection scenario}, we followed the processing settings in aforementioned scenarios and then applied different schemes (i.e., DP and attacks) into the FL pipeline. By default, all of the models were trained using SGD with a momentum of 0.9. We set the local training epoch to 5 and conducted 50 federated rounds to guarantee convergence. In addition, the hyper-parameter $z$ and $E\%$, which represent the number of initialized noises and the channel percentage of BN layers, were respectively set to 200 and 50\%. Finally, we repeated all the experiments 3 times and took the average of them as the reported results.

\begin{table}[]
\centering
\caption{Results on the category-imbalance scenario with CIFAR-10. Each \textit{"Type"} represents a unique data distribution.}
\label{tab:cifar_vgg-res}
% \vskip 0.15in
\begin{tabular}{@{}ccccccc|cccccc@{}}
\toprule
\multirow{2}{*}{\textbf{Method}}    & \multicolumn{6}{c}{\textbf{VGG-11}} &  \multicolumn{6}{c}{\textbf{ResNet-18}}   \\ \cline{2-13}
   & \textit{Type1}          & \textit{Type2}          & \textit{Type3}          & \textit{Type4}          & \textit{Type5} & \textit{Avg}& \textit{Type1}          & \textit{Type2}          & \textit{Type3}          & \textit{Type4}          & \textit{Type5}&\textit{Avg}  \\ \hline
FedAvg    & 38.65          & 40.35          & 21.25          & 32.55          & 37.05 & 33.97 & 28.60          & 35.10          & 25.30          & 32.00          & 30.45  &30.29        \\
Local     & 81.50          & 81.37          & 69.38          & 81.44          & 83.11 & 79.36 & 74.00          & 70.40          & 63.37          & 74.16          & 76.09   &71.60     \\
LG-Net    & 83.36          & 85.94          & 76.87          & 85.43          & 84.99& 83.32 & 80.43          & 79.65          & 68.05          & 79.71          & 82.15    &78.00      \\
K-Cluster & 88.60          & 92.55          & 79.45          & 91.20          & 91.35  & 88.63 & 86.40          & 86.55          & 74.80          & 55.95          & 58.25  &72.39     \\
\textbf{Ours}      & \textbf{89.70} & \textbf{93.05} & \textbf{89.50} & \textbf{91.40} & \textbf{91.75} & \textbf{91.08}& \textbf{87.80} & \textbf{88.25} & \textbf{76.60} & \textbf{87.00} & \textbf{87.40} & \textbf{85.41} \\ \bottomrule
\end{tabular}
% \vskip -0.1in
\end{table}

% \usepackage{booktabs}
% \begin{table}[]
% \centering
% \caption{CIFAR-10 results for the baseline and different personalization methods on ResNet-18.}
% \label{tab:cifar_res18}
% % \vskip 0.15in
% \begin{tabular}{@{}cccccc@{}}
% \toprule
% \textbf{Method} & \textit{Type1}          & \textit{Type2}          & \textit{Type3}          & \textit{Type4}          & \textit{Type5}          \\ \midrule
% FedAvg    & 28.60          & 35.10          & 25.30          & 32.00          & 30.45          \\
% Local     & 74.00          & 70.40          & 63.37          & 74.16          & 76.09          \\
% LG-Net    & 80.43          & 79.65          & 68.05          & 79.71          & 82.15          \\
% K-Cluster & 86.40          & 86.55          & 74.80          & 55.95          & 58.25          \\
% Ours      & \textbf{87.80} & \textbf{88.25} & \textbf{76.60} & \textbf{87.00} & \textbf{87.40} \\ \bottomrule
% \end{tabular}
% % \vskip -0.1in
% \end{table}

\subsection{Results on the Category-imbalance Scenario}
\label{sec:category-imbalance}

% \bingyan{lack: convergence line; HAR results; privacy results; model replacement; }

% \textbf{Accuracy on different models.}
As we stated in the experimental settings, there are 5 types of distributions and we test the average performance on each type. Table \ref{tab:cifar_vgg-res} illustrates the results on VGG-11 and ResNet-18. We can clearly see that the proposed DistFL consistently achieves higher accuracy than other methods across all of the types. On average, the improvement is over 2\% for VGG-11 and 7\% for ResNet-18 compared to other methods. This validates that by distribution-aware FL, we can greatly boost the performance for the category-imbalance scenario. Besides, it is worth noting that: (1) The \textit{FedAvg} baseline is worse than the \textit{Local} training, which suggests that the typical FL algorithm fails to cope with this non-iid setting and it is even not necessary to conduct federated learning under this condition; (2) The overall performance on VGG-11 outperforms ResNet-18, demonstrating that the skip connections existed in ResNets may harm the FL process when each client owns images with different categories. Despite the interference, DistFL still exceeds other methods by a large margin, which further confirms the effectiveness of our approach.

% \textbf{Convergence analysis.}
% Besides the final accuracy, we further conduct a convergence analysis by recording the intermediate accuracy output of each FL round. Figure \ref{xx} (\ref{xx}) shows the results on different methods. Note that we do not involve the \textit{LG-Net} method in the figure since it starts from a federated model, which is not comparable to other methods based on a randomly initialized model. From the figure, we can observe that: \bingyan{add according to figure? maybe not need}

\begin{table}[]
\centering
\caption{Results on the environment-difference scenario with Office-Home. Here \textit{Art, Clipart, Product, Real-World} are the four data domains under different environments.}
\label{tab:office_res50}
% \vskip 0.15in
\begin{tabular}{@{}cccccc@{}}
\toprule
\textbf{Method} & \textit{Art}           & \textit{Clipart}        & \textit{Product}        & \textit{Real-World} & \textit{Avg}     \\ \midrule
FedAvg    & 78.35          & 71.82          & 89.52          & 83.36   & 80.76       \\
Local     & 54.06          & 54.30          & 79.79          & 75.84  &  62.72      \\
LG-Net    & 75.91          & 74.18          & 89.06          & 84.59   & 80.94       \\
K-Cluster & 74.67          & 77.09          & 89.63          & 84.39    &80.46      \\
\textbf{Ours}      & \textbf{79.49} & \textbf{78.62} & \textbf{93.01} & \textbf{87.73} & \textbf{83.71}\\ \bottomrule
\end{tabular}
% \vskip -0.1in
\end{table}

\begin{table}[]
\centering
\caption{The accuracy (\%) achieved by different methods on HAR-ADL. Note that this dataset is collected from real world users, which can cover both the category-imbalance and environment-difference scenarios.}
\label{tab:har}
\begin{tabular}{@{}cccccccc@{}}
\toprule
\textbf{Client} & \multirow{2}{*}{\textbf{FedAvg}} & \multirow{2}{*}{\textbf{Local}} & \multirow{2}{*}{\textbf{LG-Net}} & \textbf{K-Cluster} & \textbf{K-Cluster} & \textbf{K-Cluster} & \multirow{2}{*}{\textbf{Ours}} \\
\textbf{Index} &  &  &  & \textbf{(K=2)} & \textbf{(K=3)} & \textbf{(K=4)} & \\  \midrule
1                      & 96.55           & 83.91          & 97.70           & 83.91                   & 100.00                  & 98.85                   & 100.00           \\
2                      &      100.00           &     38.16           &     100.00            &      88.16                   &             100.00            &           100.00              &       100.00        \\
3                      &        58.14         &        91.86        &        75.58         &   80.23                      &         100.00                &           96.51              &     96.51          \\
4                      &         78.95        &         30.26       &        94.74         &    94.74                     &          97.37               &         97.37                &     97.37          \\
5                      &     100.00            &       96.34         &       100.00          &   70.73                      &           98.78              &          98.78               &    98.78           \\
6                      &         89.61        &       76.62         &        98.70         &    92.21                     &           98.70              &             98.70            &      98.70         \\
7                      &      67.61           &      81.69          &        53.52         &     46.48                    &            57.75             &           54.93              &      67.61         \\
8                      &        68.06         &       68.06         &        34.72         &  68.06                       &            52.78             &           38.89              &       72.22        \\
9                      &        100.00         &        87.34        &        97.47         &  98.73                       &        97.47                 &          100.00               &    100.00           \\ 
10                      &         100.00        &        71.25        &      97.50           &   80.00                      &         80.00                &         97.50                &    97.50           \\
11                      &       98.78          &        93.90        &         75.61        &     80.49                    &             98.78            &           85.37              &      95.12         \\
12                      &      65.22           &       45.65         &         86.96        &    79.35                     &            100.00             &              90.22           &     97.83          \\
13                      &        83.33         &       87.78         &        97.78         &   98.89                      &           98.89              &            97.78             &       98.89        \\
14                      &        56.86         &        98.04        &       58.82          &    66.67                     &           100.00              &         59.80                &     100.00          \\
15                      &       91.36          &       86.42         &       85.19          &   98.77                      &            96.30             &            83.95             &     98.96          \\
16                      &        95.83         &        91.67        &          100.00       &  97.92                       &            88.54             &            100.00             &       98.96        \\
17                      &        96.94         &         94.90       &         96.94        &  96.94                       &             96.94            &        94.90                 &      98.98         \\
18                      &         90.63        &         90.63       &          98.96       &  98.96                       &             100.00            &           100.00              &       100.00        \\
19                      &          100.00       &        21.67        &        100.00         &   100.00                       &         100.00                 &         100.00                 &    100.00            \\
20                      &        98.96         &         89.58       &         96.88        &   98.96                      &           96.88              &          100.00               &      98.96         \\ \midrule
Avg                      &      86.84           &        76.29        &        87.35         &             86.01            &           92.96              &         89.68                &  \textbf{95.82}             \\

\bottomrule
\end{tabular}
\end{table}

\subsection{Results on the Environment-difference Scenario}
\label{sec:environment-difference}
\textbf{Performance on Office-Home.}
Here Office-Home is used to simulate the environment-difference scenario and each domain in the dataset is considered as a type of distribution. We use ResNet-50 as the backbone since it is commonly evaluated on this dataset. We average the achieved accuracy of each domain/type and report the domain performance in Table \ref{tab:office_res50}. Obviously, we can see that DistFL shows superiority in all domains with up to 3.38\% average improvement (i.e., on the \textit{Product} domain), demonstrating its applicability when the environments of clients are diverse. In addition, there are two interesting observations: (1) Different from the category-imbalance scenario, in this scenario the \textit{FedAvg} baseline can significantly outperform the \textit{Local} training, which means that we can benefit from FL although the environment of client data is different; (2) The state-of-the-art method \textit{K-Cluster} might be not very applicable to this setting considering that the achieved average accuracy is even lower than the \textit{FedAvg} baseline. We believe this is because the environment difference is hard to be distinguished only using the inference loss calculated in this method, thus leading to wrong clustering results.

\textbf{Performance on HAR-ADL.}
HAR-ADL is a human activity recognition (HAR) dataset whose data are collected from different real-world users. Therefore, the distribution is typically non-iid, including both the category-imbalance and environment-difference situations. Different from the Office-Home dataset that has specific distribution types, it is hard to estimate the detailed distribution information of HAR-ADL, which poses a higher demand to implement clustering automatically. Therefore, for the \textit{K-cluster} method, we manually try several hyper-parameter settings (i.e., K=2,3,4) and record their corresponding results. Table \ref{tab:har} demonstrates the overall performance with different methods. We can observe that: (1) On average, our DistFL can outperform other methods by a large margin, which suggests that distribution-aware FL does benefit the real-world user scenario; (2) For \textit{K-cluster}, the performance varies significantly given different hyper-parameter settings. This demonstrates that \textit{K-cluster} is largely dependent on the accurate manual setting, making it hard to be used in real-world applications since we cannot give the suitable manual setting before FL; (3) Besides the cluster-based method, other personalization methods can only reach an accuracy of 87.35\%, which is roughly 10\% lower than our approach. In other words, these methods fail to achieve good performance when the category-imbalance and environment-difference situations happen simultaneously.

\begin{figure}[t]
% \vskip 0.2in
\centering
\includegraphics[width=0.5\columnwidth]{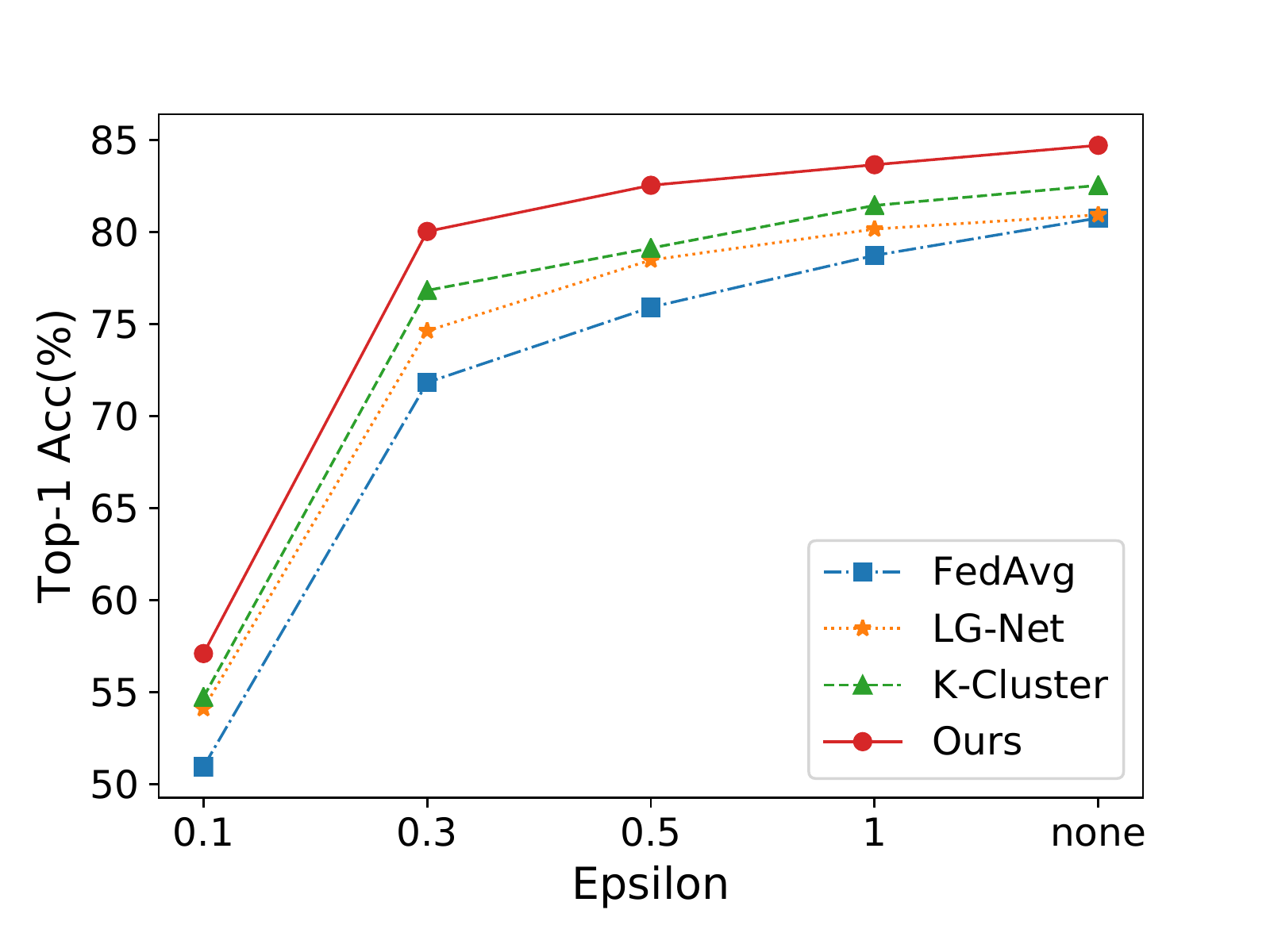} 
% \vskip -0.1in
\caption{Results on the privacy-protection scenario. Here we test the effectiveness of different methods by enforcing different privacy budgets (i.e., epsilon).}
\label{fig:privacy_analy}
% \vskip -0.2in
\end{figure}

\begin{figure}[t]
% \vskip 0.2in
\centering
\includegraphics[width=0.5\columnwidth]{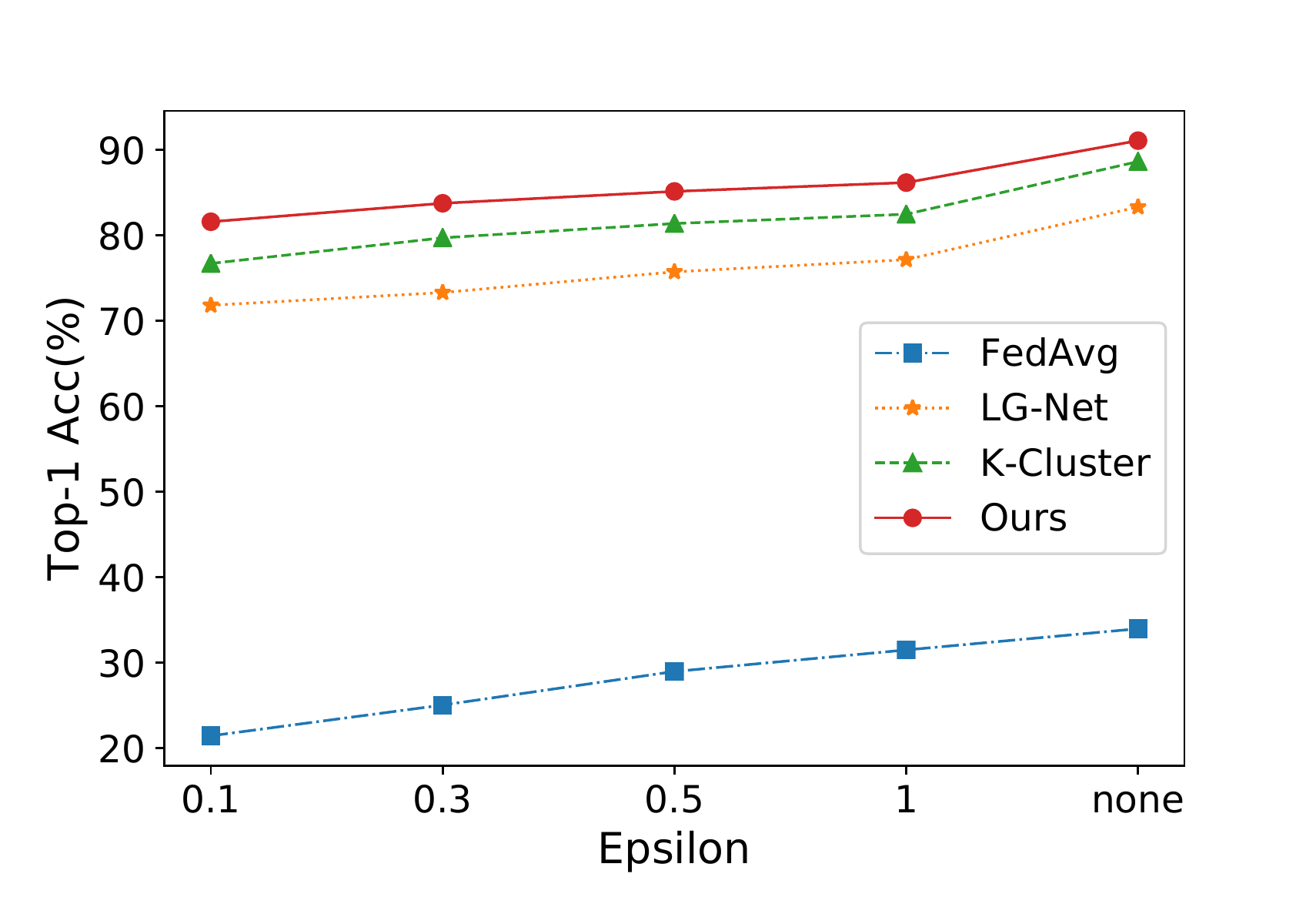} 
% \vskip -0.1in
\caption{\new{Results on the mixed scenario (i.e., category-imbalance and privacy-protection). Here we test the effectiveness of different methods by enforcing different privacy budgets (i.e., epsilon).}}
\label{fig:privacy_category}
% \vskip -0.2in
\end{figure}

\subsection{Results on the Privacy-protection Scenario}
\label{sec:privacy-enhance}
We simulate this scenario by adding differential privacy (DP) based noise to the local models for stronger privacy guarantees. The intensity of the noise is controlled by the privacy budget $\epsilon$. The smaller $\epsilon$ is, the larger noises are added, which indicates stronger defense to the inference attack. Under this privacy-protection scenario, we explore whether our approach is still effective in clustering models with different distributions.

Here we use the results on \textit{Office-Home \& ResNet-50} as an example and apply different $\epsilon$ values to local models to control the degree of privacy protection. As shown in Figure \ref{fig:privacy_analy}, the average accuracy of four domains with several methods is reported under the different $\epsilon$. From the figure, we can draw the following conclusions: (1) a larger $\epsilon$ (weaker privacy guarantee) usually leads to higher accuracy, which is intuitive since it introduces smaller noises to the models; (2) as we strengthen the degree of privacy protection (decrease $\epsilon$), the performance on all of the methods is degraded. In particular, when we decrease the $\epsilon$ from 0.3 to 0.1, almost all of the methods drop significantly and become nearly invalid; (3) our approach can achieve consistently higher accuracy regardless of the intensity of $\epsilon$, which further indicates the effectiveness of DistFL.

\new{\textbf{Performance on a mixed scenario.}}
\new{To explore whether our framework is effective in the mixed scenario, we further conduct an experiment on both the category-imbalance and privacy-protection scenario. Concretely, we employ the simulation setting in the category-imbalance scenario and incorporate different privacy budgets (i.e., $\epsilon$) to local models to enable privacy protection. Here we use the VGG-11 as the backbone and the average performance among these clients is recorded for comparison. Figure \ref{fig:privacy_category} demonstrates the results. We can observe that the proposed framework can achieve better accuracy performance than others for all of the privacy budgets (e.g., 4.88\% improvement when $\epsilon=0.1$), validating its usefulness in the mixed scenario. Besides, the baseline \textit{FedAvg} performs significantly worse than other methods, which keeps the same observation as the category-imbalance scenario where the non-iid situation largely degrades the performance of traditional aggregation used in \textit{FedAvg}.
}

\begin{table*}[]
\centering
\caption{Results on the attack-injection scenario. Here is the defense performance to the label flipping attack on VGG-11. \textit{ACC} is the testing accuracy on the final aggregated model and \textit{ASR} represents the attack success rate.}
\label{tab:defense-data-poison}
% \vskip 0.15in
\begin{tabular}{@{}c|cc|cc|cc|cc|cc@{}}
\toprule
\multirow{2}{*}{\textbf{Method}} & \multicolumn{2}{c|}{\textbf{5}}     & \multicolumn{2}{c|}{\textbf{10}}    & \multicolumn{2}{c|}{\textbf{15}}    & \multicolumn{2}{c|}{\textbf{20}}    & \multicolumn{2}{c}{\textbf{40}}    \\ \cmidrule(l){2-11} 
                                 & \textit{ACC(\%)} & \textit{ASR(\%)} & \textit{ACC(\%)} & \textit{ASR(\%)} & \textit{ACC(\%)} & \textit{ASR(\%)} & \textit{ACC(\%)} & \textit{ASR(\%)} & \textit{ACC(\%)} & \textit{ASR(\%)} \\ \midrule
FedAvg                           & 67.89            & 5.18             & 59.33            & 19.57            & 59.33            & 40.33            & 40.85            & 52.87            & 36.61            & 70.08            \\
Credit                           & 69.48            & 3.82             & 62.22            & 14.62            & 49.44            & 37.65            & 42.12            & 51.92            & 36.45            & 70.45            \\
IRLS                             & 70.24            & 3.03             & 69.93            & 3.77             & 68.14            & 4.19             & 68.06            & 4.28             & 37.09            & 65.87            \\
GRA                              & 67.40            & 4.23             & 63.26            & 4.37             & 60.65            & 7.13             & 58.14            & 8.69             & 57.25            & 10.71            \\
\textbf{Ours}                             & \textbf{70.71}   & \textbf{2.82}    & \textbf{70.01}   & \textbf{2.72}    & \textbf{69.65}   & \textbf{2.93}    & \textbf{69.47}   & \textbf{2.98}    & \textbf{65.30}   & \textbf{3.38}    \\ \bottomrule
\end{tabular}
% \vskip -0.1in
\end{table*}

\begin{table}[]
\centering
\caption{Results on the attack-injection scenario. Here is the defense performance to the model replacement attack on ResNet-18.}
\label{tab:defense-model}
\begin{tabular}{@{}cccccc@{}}
\toprule
\textbf{Metric} & \textbf{FedAvg} & \textbf{Credit} & \textbf{IRLS} & \textbf{GRA} & \textbf{Ours} \\ \midrule
\textit{ACC(\%)}               & 31.84           & 72.68           & 72.63         & 71.53        & \textbf{72.69}         \\
\textit{ASR(\%)}               & 56.15           & 2.26            & 2.35          & 2.48         & \textbf{2.22}          \\ \bottomrule
\end{tabular}
\end{table}

\subsection{Results on the Attack-injection Scenario}
\label{sec:attack-inject}

Although DP can be introduced to provide stronger privacy protection as stated in the last subsection, it is impossible to defend the poison attacks, which inject malicious behaviors on the internal models. In this scenario, we demonstrate how to inject poison attacks and the corresponding defense performance with different methods.

\textbf{Threat models.} 
We construct two threat models by implementing two types of poison attacks. The first one is the label flipping attack \cite{tolpegin2020data}, where the labels of honest training examples of one class are flipped to another class while the features of the data are kept unchanged. Another is the model replacement attack \cite{bagdasaryan2020backdoor}, which accomplishes the attack by replacing the local model with elaborate weights to mislead FL. Note that this replacement attack will generate an unusable local model that cannot be processed using our extraction techniques. Thus, we resort to a \textit{pre-aggregation} scheme since the aggregated model is always a usable model. 

\textbf{Defense on these attacks.} 
We use two metrics to evaluate the defense performance: \textit{testing accuracy on the final aggregated model} and \textit{ASR}. The first metric is to examine whether the influence of malicious clients can be removed. \textit{ASR} represents the attack success rate, which is higher if the misleading accuracy on the targeted input increases. \new{In our scenario, \textit{ASR} is defined as the success probability of perturbing a specific image class to a target class, which can be calculated by 
\begin{equation}
    ASR=\frac{\sum_{i=1}^{|D_{s}|} \mathbb{I}(M_{agg}(x_i)==y_{target})}{|D_{s}|}
\end{equation}
Here $|D_{s}|$ is the number of the specific image class and $M_{agg}$ is the aggregated model.  $y_{target}$ is the label of the target class and $x_i$ is the $i_{th}$ sample of the specific class. $\mathbb{I}$ is an indicator function. If the attack includes multiple classes, we average their ASR as the final result.}

Table \ref{tab:defense-data-poison} shows the defense results to the label flipping attack on VGG-11. Here label 0-3 and label 4-7 are mutually flipped to conduct the misleading training. We respectively implement the attack on 5, 10, 15, 20 and 40 clients of the simulated 50 clients, to test the defense performance on different attack scales. \new{Note that under the label flipping attack, DistFL will generate two aggregated models for the normal and abnormal cluster respectively, and we use the model in the normal cluster as the aggregated model because this model is the decent model to be deployed for mobile users (another one is just an attack model).} From the table, we can observe that although other methods such as \textit{IRLS} and \textit{GRA} can resist the attack, they fail to accomplish the defense goal as the number of attacked clients grows, especially when the attack scale is large enough (e.g., 40 malicious clients in our simulation). \new{Different from them, our approach has the capability to achieve good defense performance even though the attack scale is large (e.g., 40/50 users are attackers and they flip their data labels). This is because in DistFL, the malicious clients are solely and accurately clustered, getting rid of their negative influence on the final aggregated model of normal clients.} Besides, our accurate clustering also benefits more to the testing accuracy compared with other methods since no malicious behavior can be introduced to disturb the FL process in the cluster of the normal clients. 

For the model replacement attack, we also use the same label setting as the flipping attack and conduct the misleading training. After training, we replace the model weights using techniques stated in \cite{bagdasaryan2020backdoor} to achieve a stronger attack with only one malicious client. As shown in Table \ref{tab:defense-model}, the attack can significantly harm traditional FL (FedAvg). However, all of the defense methods are able to defend it due to its over weird weights, which can be easily detected and removed. Compared to other methods, our approach performs slightly better in an efficient and automated way.

\begin{table}[]
\centering
\caption{\new{The accuracy (\%) achieved by different methods on the real-world dataset collected from a mobile app.}}
\label{tab:real}
\begin{tabular}{@{}cccccccc@{}}
\toprule
\textbf{Client} & \multirow{2}{*}{\textbf{FedAvg}} & \multirow{2}{*}{\textbf{Local}} & \multirow{2}{*}{\textbf{LG-Net}} & \textbf{K-Cluster} & \textbf{K-Cluster} & \textbf{K-Cluster} & \multirow{2}{*}{\textbf{Ours}} \\
\textbf{Index} &  &  &  & \textbf{(K=2)} & \textbf{(K=3)} & \textbf{(K=4)} & \\  \midrule
1                      & 75.63           & 50.00          & 86.25          & 85.63                   & 86.25                  & 83.13                   & 87.50           \\
2                      &      80.94           &     77.05           &     92.62            &      90.33                   &             90.12            &           90.53              &       92.83        \\
3                      &        97.06         &        42.65        &        95.59         &   96.32                      &         95.59                &           94.85              &     97.06          \\
4                      &         73.96        &         70.83       &        89.58         &    83.33                     &          89.58               &         88.54                &     89.58          \\
5                      &     78.41            &       80.11         &       86.43          &   81.82                      &           85.23              &          81.82               &    86.43           \\
6                      &         83.82        &       93.38         &        94.12         &    94.12                     &           91.91              &             91.18            &      95.59         \\
7                      &      83.33           &      83.33          &        89.46         &     90.69                    &            90.69             &           90.69              &      91.38         \\
8                      &        41.67         &       35.42         &        79.17         &  75.35                       &            75.00             &           75.69              &       80.21        \\
9                      &        68.45         &        53.81        &        81.55         &  70.83                       &        77.38                 &          77.98               &    85.12           \\ 
10                      &         75.40        &        81.85        &      88.71           &   93.15                      &         91.53                &         89.92                &    93.15           \\
11                      &       96.02          &        89.77        &         100.00        &     98.30                    &             98.86            &           98.30              &      100.00         \\
12                      &      85.00           &       82.50         &         91.75        &    90.00                     &            90.00             &              85.00           &     91.75          \\
13                      &        83.93         &       50.00         &        91.07         &   83.93                      &           87.50              &            82.14             &       92.64        \\
14                      &        62.50         &        71.20        &       77.72          &    75.54                     &           75.54              &         76.63                &     78.26          \\
15                      &       80.56          &       45.83         &       87.50          &   87.50                      &            81.02             &            87.50             &     87.50          \\
16                      &        87.69         &        53.85        &          90.79       &  87.69                       &            87.69             &            89.90             &       92.71        \\
17                      &        88.54         &         67.81       &         91.04        &  82.31                       &             91.04            &        91.04                 &      92.38         \\
18                      &         93.75        &         68.75       &          91.88       &  96.88                       &             93.75            &           93.75              &       93.75        \\
19                      &          83.17       &        57.07        &        90.22         &   86.61                       &         86.61                 &         89.50                 &    92.93            \\
20                      &        79.51         &         68.40       &         92.93        &   90.63                      &           92.93              &          94.64               &      94.64         \\ \midrule
Avg                      &      79.97           &        66.18        &        89.42         &             87.05            &           87.91              &         87.64                &  \textbf{90.77}             \\

\bottomrule
\end{tabular}
\end{table}

\new{\subsection{Results on the Real-world Scenario}
\label{sec:real}}
\new{Aforementioned scenarios are based on the simulation setting with public datasets. However, the data distribution in the real-world scenario is more complex and even has no specific pattern. To further test the generality of the proposed approach, we conduct an experiment on a human activity recognition (HAR) dataset collected from a mobile app to represent the real-world scenario \cite{li2021meta}, where the data distribution is not clear as our simulated scenarios. Specifically, an Android app is first developed for activity signal logging and 6 types of activities, including \textit{Walking, Biking, (walking) Upstairs, (walking) Downstairs, Running and Taking Bus/Taxi}, are collected to form the dataset. Similar to the setting used in HAR-ADL, we select 20 users as 20 clients ( 17,382 training samples and 4,381 testing samples) and employ a simple 2-layer CNN to conduct the experiment. }

\new{Table \ref{tab:real} summaries the results. From the table, we can obtain the following observations: (1) It is obvious that the proposed approach achieves the best performance on average, which suggests the effectiveness of DistFL under the real-world scenario; (2) Unlike the results of the HAR-ADL dataset where \textit{LG-Net} performs worse than \textit{K-Cluster}, in this context, \textit{LG-Net} can exceed \textit{K-Cluster} by up to 2.37\% average accuracy. We believe this is because in this scenario we own more training samples than HAR-ADL, which will significantly mitigate the overfitting problem to the local fine-tuning process existed in \textit{LG-Net}. However, \textit{LG-Net} needs more computation cost than others since it includes a \textit{FedAvg} process plus a fine-tuning process, which increases the burden on the resource-constrained mobile devices; (3) The hyper-parameter $K$ in \textit{K-Cluster} is not very sensitive in this scenario since the final accuracy of different $K$ values is similar. However, all of the hyper-parameter settings for  \textit{K-Cluster} perform worse than DistFL, which indicates that our clustering result is more precise and thus benefits the final performance.
}

\begin{figure*}[t]
% \vskip 0.2in
\centering
\includegraphics[width=1\columnwidth]{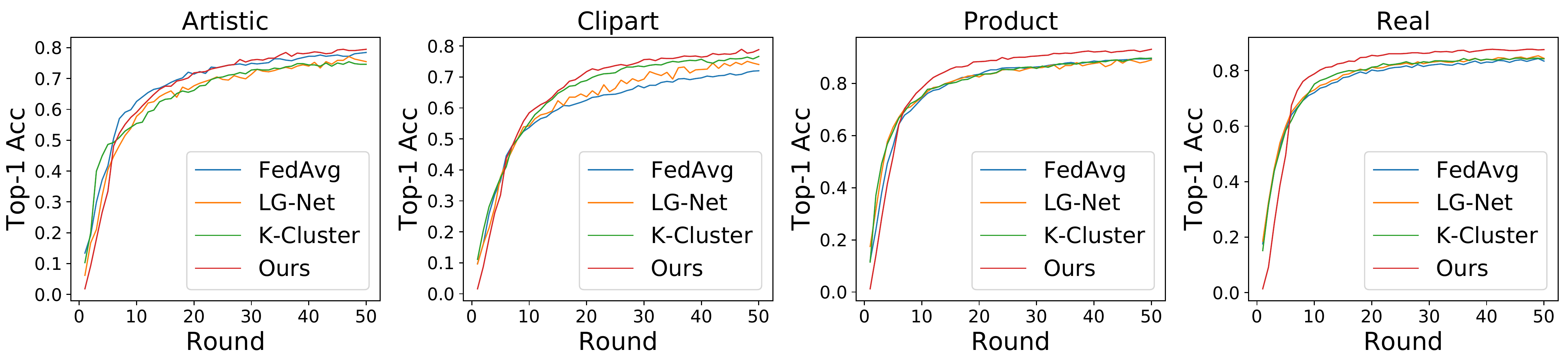} 
% \vskip -0.1in
\caption{\new{Convergence performance of the environment-difference scenario with Office-Home.}}
\label{fig:converge}
% \vskip -0.2in
\end{figure*}

\new{\subsection{Convergence Performance}
\label{sec:convergence}}
\new{To better observe the training process of different methods, we record their corresponding testing accuracy in each FL round and plot the convergence lines. Here we conduct the experiment on the environment-difference scenario with Office-Home and for each domain, we randomly pick out one device as an example. As shown in Figure \ref{fig:converge}, all the methods can reach convergence within 50 FL rounds and DistFL achieves better performance compared to others. In addition, it is worth noting that: (1) For the \textit{Product} and the \textit{Real-World} domain, the convergence speed is significantly faster than the other two domains. Specifically, we can obtain desirable performance at roughly 30 FL rounds, which suggests that the two domain tasks are easier to train; (2) Other methods behave unstably on different domains. For example, \textit{FedAvg} can achieve comparable convergence speed as ours on the \textit{Artistic} domain while failing to perform well on the \textit{Clipart} domain. Different from them, DistFL can keep consistent superiority under any domain situation.
}

\begin{figure}[t]
% \vskip 0.2in
\centering
\includegraphics[width=0.8\columnwidth]{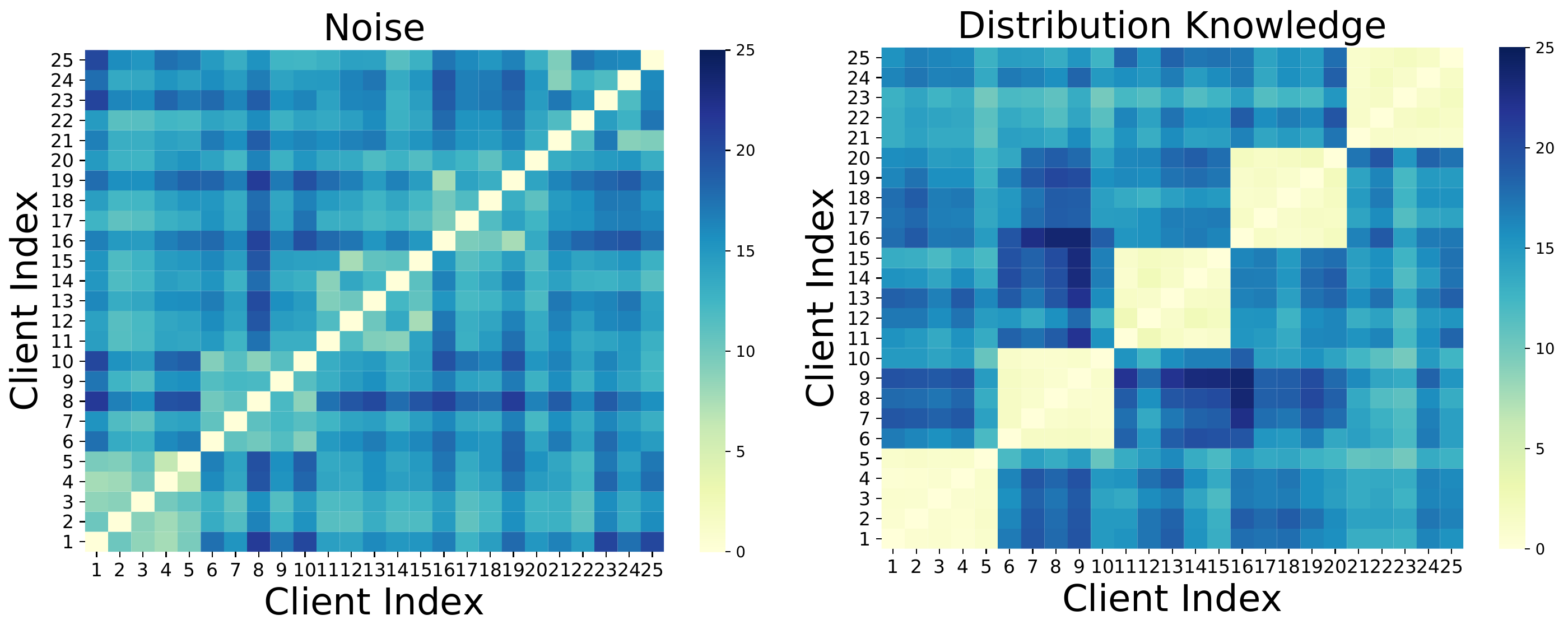} 
% \vskip -0.1in
\caption{Visualization of the similarity matrix generated by noises (left) and our distribution knowledge (right). Lighter color indicates higher similarity. We can clearly see that using the distribution-based matrix can provide better guidance to the clustering process.}
\label{fig:noise_konwledge}
% \vskip -0.2in
\end{figure}

\subsection{Effectiveness and Efficiency Analysis}
\label{sec:detailed_analy}
This subsection analyzes different modules and parameters used in DistFL to provide readers a better understanding of the effectiveness and efficiency of our framework. All of the analyses, unless otherwise specified, are based on the setting of the \textit{category-imbalance scenario}.

\begin{figure}[t]
% \vskip 0.2in
\centering
\includegraphics[width=0.9\columnwidth]{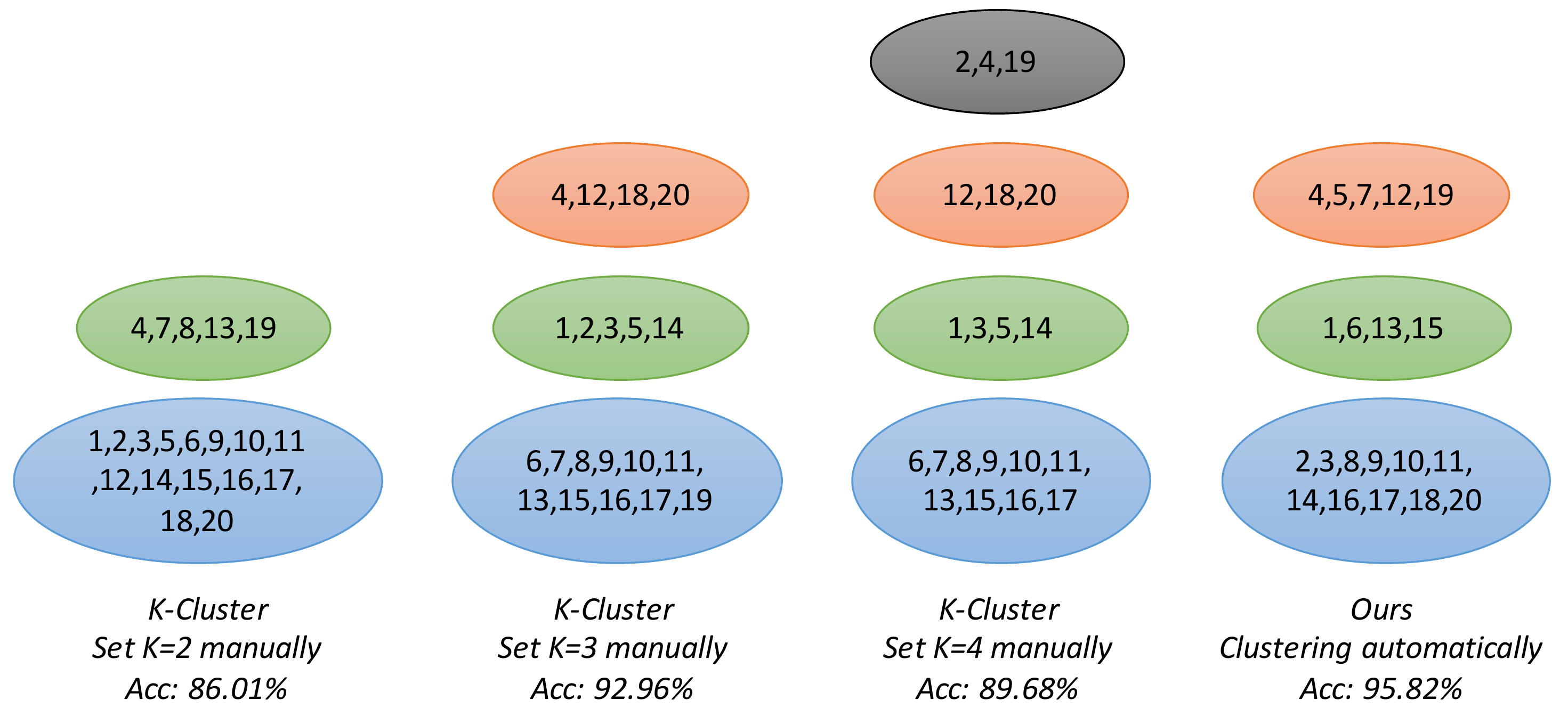} 
% \vskip -0.1in
\caption{Illustration of the concrete cluster information on HAR-ADL, which owns data collected from 20 different users. In the figure we use the client index to represent each user and allocate them into different clusters based on the K-Cluster method and our method. Compared to K-Cluster, our distribution-aware clustering can achieve a better accuracy without manual efforts.}
\label{fig:cluster_information}
% \vskip -0.2in
\end{figure}

\textbf{Effectiveness of the extracted knowledge.}
We conduct two control experiments to verify the effectiveness of the distribution knowledge, which are summarized as follows.
\begin{itemize}
    \item \textit{Noise vs. Distribution knowledge.}
    In this experiment, we respectively feed some noises and the generated distribution knowledge into models to obtain the similarity matrix. Here we randomly picked out 25 clients and every 5 with continuous indexes come from the same distribution. Figure \ref{fig:noise_konwledge} depicts the results, where the lighter color denotes higher similarity. Obviously, it is hard to achieve clustering from the noise-based matrix since we cannot distinguish suitable clients from others given a target client. Different from the noises, our distribution knowledge is able to obtain a desirable similarity matrix, where every 5 clients of the same distribution are clearly grouped, confirming the usefulness of the generated knowledge.
    
    \item \textit{Loss-based clustering vs. Distribution-based clustering.}
    In this experiment, we compare the concrete cluster results with the loss-based clustering (K-Cluster \cite{ghosh2020efficient}) and our distribution-based clustering. Here we use the HAR-ADL dataset as an example. As shown in Figure \ref{fig:cluster_information}, the 20 client models are distributed to different clusters denoted in different colors. We can conclude that the loss-based scheme obtains an inappropriate clustering result despite using various parameter settings since all of their final accuracy are significantly lower than ours. Besides, among these manual settings, partitioning models into 3 clusters can achieve the best performance, which is consistent to our approach. However, we would like to point out that our approach is automated and the concrete model index in each cluster is different from the loss-based scheme. That is why our approach can outperform others.
\end{itemize}

\textbf{Parameter selection for better efficiency.}
In our framework, the computation costs come from two aspects: (1) the amount of knowledge we extract; (2) the number of extraction target in a model. Here we implement some experiments with different parameters to explore the best choice for efficiency.

\begin{figure}[t]
% \vskip 0.2in
\centering
\includegraphics[width=1\columnwidth]{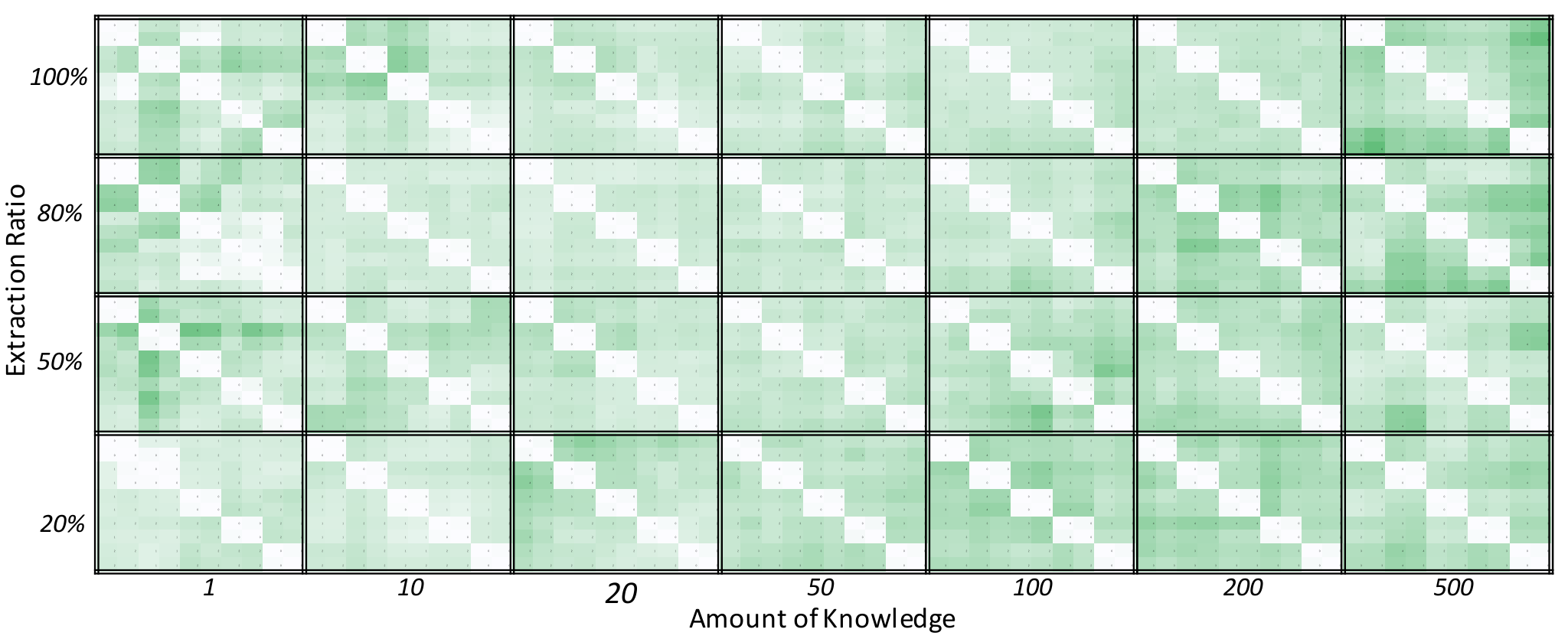} 
% \vskip -0.1in
\caption{Influence of the extraction ratio and the amount of knowledge on the similarity matrix. Here we randomly sample 10 clients and every 2 continuous clients come from a similar distribution. Thus, for each matrix, every four blocks on the diagonal indicate the models with similar distributions. Lighter color means higher similarity.}
\label{fig:detailed_analysis}
% \vskip -0.2in
\end{figure}

For the first aspect, as stated in Section \ref{sec:knowledge_group}, the distribution knowledge can be denoted as $K_{target}=\{k_{target}^1,k_{target}^2,\\...,k_{target}^z\}$, where the amount of knowledge is reflected by the parameter $z$. Thus we explore the influence of $z$ on the clustering performance. Specifically, the value of $z$ is set to 1, 10, 20, 50, 100, 200 and 500 and we respectively use these values to generate the corresponding knowledge and the similarity matrix of 10 randomly sampled clients, where every two continuous clients come from the same distribution. We plot the similarity matrix of each parameter pair (i.e., ($z,E\%$), we will analyze the influence of $E\%$ in the following part) and for each matrix, every four blocks on the diagonal indicate the models with similar distributions. Here lighter color means higher similarity. As shown in Figure \ref{fig:detailed_analysis}, the clustering performance becomes more significant when we increase the amount of generated knowledge (i.e., the value of $z$). Generally, 100 ``distribution knowledge'' (i.e., $z=100$) is enough to achieve good clustering regardless of the extraction ratio $E\%$.

Besides, deciding the number of extraction target is also necessary for efficiency since there are many BN layers in a modern DNN. In DistFL, we propose to only optimize the important BN channels during the extraction. To study the impact of different extraction ratios (i.e., $E\%$) of BN layers, we respectively select 20\%, 50\%, 80\% and 100\% BN channels of each BN layer to conduct our pipeline. The performance is also measured by the similarity matrix as illustrated in Figure \ref{fig:detailed_analysis}. From the figure, we can clearly see that it is not necessary to optimize the whole BN channels, especially when the amount of the extracted knowledge is enough (e.g., 100). This demonstrates that we can accomplish the extraction process efficiently by employing fewer BN channels such as 50\% or even 20\% when we have enough quantity of the distribution knowledge.

\section{Discussions}
\label{sec:discussions}
One limitation of the paper is the lack of experiments on real-world applications. Although we have simulated different mobile scenarios to validate the effectiveness of our approach, the real-world study is also an important aspect worth exploring further. In our future work, we will look for ways to evaluate DistFL in real-world applications. Besides, this paper mainly focuses on computer vision tasks and human activity recognition tasks. It is interesting to apply our framework for other deep learning tasks by analyzing and extracting the distribution knowledge from their models. Finally, although we have simulated the privacy-protection scenario by enforcing differential privacy (DP) to the local models, researchers may want to use some secure aggregation schemes \cite{bonawitz2017practical} to further protect the model information. However, secure aggregation requires massive computation power, which may be unacceptable to the resource-constrained clients under our mobile scenarios.

\new{Besides, we provide some findings and lessons learned from this work. First, before generating bogus data, researchers should check if the BN layer has been used in the model although it is prevalent in the current architectures. If not, the statistical information needs to be collected manually and then uploaded to the server for data generation. Second, the performance of our clustering process depends on the responses of each model, which are obtained by feeding the distribution knowledge (i.e., bogus data) into each local model. Therefore, it is necessary to generate enough bogus data to make the distribution knowledge more precise, which will then benefit the final clustering process. }

% Under this scenario, we need to incorporate our approach into the aggregation protocol and we leave this design as future work.
   
\section{Conclusion}
\label{sec:conclusion}
This paper proposes \textit{distribution-aware federated learning} to address the non-iid problem, which widely exists in mobile scenarios. We design and implement a framework named DistFL to accomplish our goal efficiently, automatically and accurately, where the distribution knowledge of federated models is extracted to distinguish and cluster different client models. Experimental results demonstrate that DistFL achieves significantly better performance on all simulated scenarios with different models compared to state-of-the-art approaches.

\begin{acks}
We would like to thank the anonymous reviewers for their valuable feedback. This work was partly supported by the National Key Research and Development Program (2017YFB1001904) and the National Natural Science Foundation of China (61772042).
\end{acks}

%%
%% The next two lines define the bibliography style to be used, and
%% the bibliography file.
\bibliographystyle{ACM-Reference-Format}
\bibliography{sample-base}

\end{document}